\documentclass[10pt,twocolumn,letterpaper]{article}

\usepackage[pagenumbers]{cvpr}

\usepackage{iftex}
\usepackage{graphicx}
\usepackage{booktabs}
\usepackage{multirow}
\usepackage{amsmath}
\usepackage{amsthm}
\newtheorem{proposition}{Proposition}
\ifLuaTeX\else\usepackage{amssymb}\fi
\usepackage{xcolor}
\graphicspath{{figures/}}

\providecommand{\bdelta}{\boldsymbol{\delta}}

\usepackage[protrusion=true,expansion=true,stretch=20,shrink=20]{microtype}
\definecolor{cvprblue}{rgb}{0.21,0.49,0.74}
\usepackage[pagebackref,breaklinks,colorlinks,allcolors=cvprblue]{hyperref}

\ifLuaTeX
  \pdfvariable suppressoptionalinfo \numexpr 1+2+4+8+32+64\relax
\else\ifPDFTeX
\fi\fi

\def\confName{CVPR}
\def\confYear{2027}

\newcommand{\ours}{UltimateFlow}

\title{\ours: Frame-Guided Event Optical Flow\\with Depth Foundation Priors}
\author{Chahat Deep Singh\\
University of Colorado Boulder\\
{\tt\small chahat.singh@colorado.edu}\\
{\small Project page: \url{https://www.praisecu.com/research/UltimateFlow}}}

\usepackage{capt-of}
\makeatletter
\renewcommand\paragraph{\@startsection{paragraph}{4}{\z@}{1.2ex \@plus .4ex \@minus .2ex}{-1em}{\normalfont\normalsize\bfseries}}
\makeatother
\begin{document}
\twocolumn[{%
\renewcommand\twocolumn[1][]{#1}%
\maketitle
\begin{center}
\vspace{-1.9em}
\includegraphics[width=0.97\textwidth]{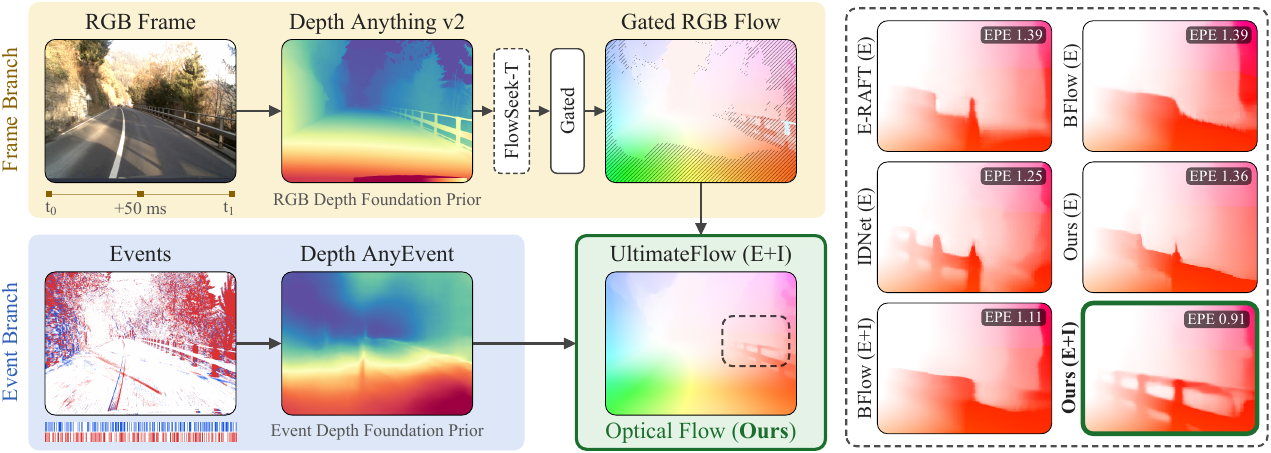}
\captionof{figure}{\textbf{\ours{} on a Novel DSEC Test Sequence.}
(Top) The frozen FlowSeek-T optical flow model, built on the Depth Anything v2 foundation model, estimates flow, gated
by a forward-backward consistency check.
(Bottom) \ours{} estimates flow from the events and their Depth AnyEvent
foundation model features, initialized and guided by the gated RGB flow as a dense input. (Right) Close-ups of the dashed region on one color scale, with each method's benchmark EPE on this sequence. Inputs: (E) events, (E+I) events and RGB frames.}
\label{fig:teaser}
\end{center}%
}]

\begin{abstract}
Event cameras capture fast motion and extreme lighting where frame cameras
fail, making event-based optical flow essential for agile robots and
autonomous driving. Since events vanish in static and untextured regions,
recent methods fuse them with RGB frames. However, current fusion methods are only marginally better than using events alone. In this paper, we introduce \ours, which rethinks how frames should be used for event-based optical flow. Rather than fusing frame features, we utilize a frozen RGB optical flow network, built on a depth foundation model, to provide an initial flow estimate and a per-pixel reliability map. A recurrent event-based estimator, built on an event depth foundation model, then refines this estimate without requiring additional labels. \ours{} achieves state-of-the-art results on the DSEC optical flow benchmark, ranking first of $39$
entries with an endpoint error (EPE) of $0.554$.
\end{abstract}

\section{Introduction}
\begin{figure}[t]
\centering
\includegraphics[width=\linewidth]{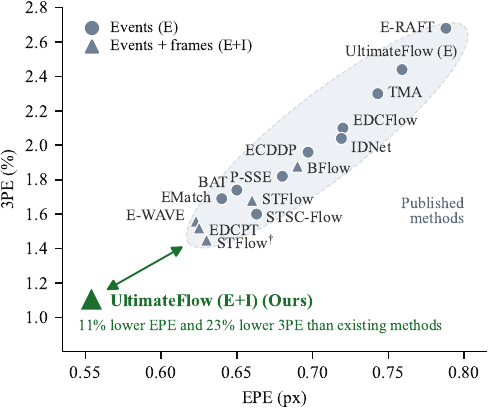}
\caption{\textbf{DSEC-Flow Test Server}: EPE against 3PE for every published method of
\cref{tab:leaderboard}. Circles: events, triangles: events and frames. $^{\dagger}$Pretrained on
synthetic event data. \ours{} (E+I) is the 5-model ensemble. Lower left is better.}
\label{fig:dsecscatter}
\end{figure}
\label{sec:intro}
Optical flow, the apparent motion of every pixel between two instants, is a basic building
block of visual odometry, autonomous driving and robot control. Frame cameras struggle in exactly
the conditions these applications face. Fast motion blurs the image, a limited dynamic range
saturates dark and bright regions, and a fixed frame rate trades bandwidth against latency. Event
cameras avoid these limits. Instead of capturing images at regular intervals, each pixel reports
brightness changes asynchronously, with microsecond resolution, a dynamic range above $120$\,dB
and no motion blur~\cite{gallego2022survey}, making them ideal for fast, low-light robotics
applications.

These advantages come with a fundamentally different signal. A pixel reports nothing unless its
brightness changes, so the signal vanishes on untextured surfaces, in momentarily static regions
and in saturated night scenes, exactly where correspondence is
hardest. Furthermore, real-world labels are scarce. DSEC-Flow~\cite{gehrig2021eraft} and
MVSEC~\cite{zhu2018mvsec} derive sparse flow from LiDAR and pose on a handful of driving and flight sequences,
and larger synthetic event datasets~\cite{li2023blinkflow} leave a gap to real sensors.

Most platforms mount a frame camera beside the event sensor, and frames record
exactly the texture that events miss. Recent methods therefore fuse the two
modalities in joint feature encoders~\cite{gehrig2024dense,stflow2025}, but the
gain has been small. On the DSEC-Flow benchmark, the best methods that combine
events and frames reach $0.62$ EPE~\cite{wu2026ewave,wan2025edcpt}, against $0.64$ for EMatch~\cite{zhang2025ematch}. Meanwhile, frame-based flow, trained on far larger datasets, has improved markedly by building on
depth foundation models~\cite{poggi2025flowseek}, and such a model now exists for events as
well~\cite{bartolomei2025dae}. How an event-based estimator should
use this frozen knowledge remains open.

A straightforward way to exploit a frozen frame-based flow model, which we call the \emph{guide},
is to blend its flow with the event-based estimate wherever the guide is
self-consistent. With all components tuned, this output-level fusion significantly improves the EPE
of our event-only model on DSEC-Flow. However, such a blend is fundamentally limited, since a single
weight is shared by all pixels and the event-based estimator never observes the guide.

In this work, we introduce \ours{} (\cref{fig:teaser}), which, instead of blending the guide at the
output, uses it as an input to the event-based estimator. The guide flow, along with its
forward--backward residual and the gate, enters the estimator as additional dense inputs, and the
recurrent refinement starts from the guide flow wherever the gate accepts it. The estimator thus
decides, per pixel, how much to trust the guide, and refines it instead of searching for the
motion from scratch. The dense inputs recover
most of the gain and the initialization the rest, especially under large motion
(\cref{sec:guidanceabl,sec:basin}). As a result, \ours{} outperforms the entire output-level fusion
stack and achieves state-of-the-art results on the DSEC-Flow benchmark (\cref{tab:leaderboard,tab:stack}, \cref{fig:dsecscatter}). Our contributions are as follows:
\begin{enumerate}
\setlength{\itemsep}{1pt}
\item \textbf{Dense Frame Guidance.} A simple and effective alternative to blending frame-based
and event-based flow at the output. A frozen
frame-based flow model guides the event-based estimator through dense inputs, gated by its own
forward--backward consistency, and initializes the recurrent refinement wherever it is reliable. We show that rotation-only registration leaves a one-sided
error band at motion boundaries that no output-level blend can remove (\cref{prop:parallax}), and
that our guided estimate halves the guide's error on its affected side. Controlled ablations show
that this outperforms every output-level fusion we could build and reveal when the initialization
matters (\cref{sec:dense,sec:guidanceabl,sec:basin}).
\item \textbf{Depth-Foundation Event Estimator.} To our knowledge, the first event-based flow
estimator conditioned on a frozen event depth foundation model, compact enough to train on a
single GPU (\cref{sec:estimator}).
\item \textbf{State of the Art.} \ours{} ranks first on DSEC-Flow, with the lowest EPE of
all entries (\cref{sec:benchmark}).
\end{enumerate}

\section{Related Work}
\label{sec:related}
\begin{figure*}[t]
\centering
\includegraphics[width=\textwidth]{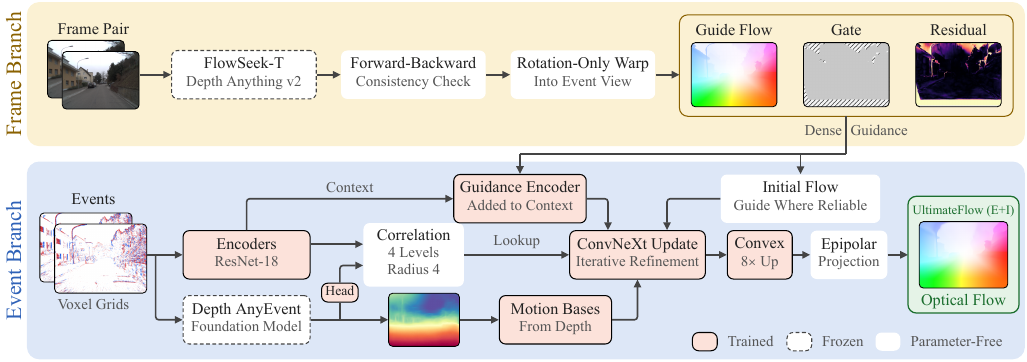}
\caption{\textbf{Architecture.} Frame branch: the frozen FlowSeek-T estimates the guide flow between the two frames in both directions, a forward--backward consistency check gives the residual and the gate, and a rotation-only warp brings all three into the event view. Event branch: a trained head feeds the Depth AnyEvent features into the correlation, and the predicted depth gives the motion bases. The dense guidance enters the context and sets the initial flow wherever the gate accepts it, and iterative ConvNeXt updates refine the flow.}
\label{fig:arch}
\end{figure*}
\paragraph{Event-Based Optical Flow.} Learning-based methods dominate event-based optical flow.
E-RAFT~\cite{gehrig2021eraft} brought recurrent all-pairs correlation to event voxel grids, and
subsequent methods explore temporal aggregation~\cite{liu2023tma}, iterative
deblurring~\cite{wu2024idnet}, continuous-time trajectories~\cite{gehrig2024dense}, state-space
encoders~\cite{psse2025} and dense matching~\cite{zhang2025ematch}, among other
designs~\cite{xu2026bat,liu2025edcflow,luo2024eemflow,hu2026stsc}. Unsupervised methods instead learn
from the alignment of motion-compensated events~\cite{zhu2019unsupervised,ye2020ecn}, and spiking
networks target energy-efficient hardware~\cite{lee2020spikeflownet}. None of these event-only methods
conditions on a depth foundation model. Our event branch builds on E-RAFT's recurrent correlation
and adds such a prior.

\paragraph{RGB and Event-Based Optical Flow.} Since DSEC provides time-aligned frames, several methods
combine both modalities. Wan~\etal~\cite{wan2022dceiflow} estimate dense flow from a single image and
events. BFlow's frame variant~\cite{gehrig2024dense}, STFlow~\cite{stflow2025},
EDCPT~\cite{wan2025edcpt} and E-WAVE~\cite{wu2026ewave} fuse frame and event features inside one
trained network, E-WAVE with features of the frozen Depth Anything v2 foundation model, while
others fuse the boundaries and motion correlation of the two
modalities~\cite{zhou2025bridge,wang2025diffabflow}. In contrast, we keep a frozen frame-based flow
model intact and pass its output and self-consistency to the event-based estimator as inputs, so
that the frame branch can be evaluated and gated independently (\cref{sec:fusionrules}).

\paragraph{Depth Priors for Optical Flow.} Monocular depth foundation models such as Depth Anything
v2~\cite{yang2024dav2} and Depth Pro~\cite{bochkovskii2025depthpro} generalize zero-shot across
scenes, which makes them attractive priors for motion estimation. FlowSeek~\cite{poggi2025flowseek} combines the Depth Anything v2
foundation model, SEA-RAFT~\cite{wang2024searaft} and motion bases for
frame-based flow, and Depth AnyEvent~\cite{bartolomei2025dae} adapts this foundation model to events.
We use the frozen DSEC fine-tune of Depth AnyEvent, so the event branch sees no frames.

\paragraph{RGB Frames as Supervision and as Input.} Frames have long supervised event-based flow,
starting with the photometric error of co-registered frames in EV-FlowNet~\cite{zhu2018evflownet},
synthetic data fills the gap in labels~\cite{li2023blinkflow,gehrig2024dense}, and a frame-based
tracker has been distilled into an event-based motion estimator~\cite{yang2026teto}. Differently
from these works, we use frames neither as targets nor in the loss. The guide's flow and its
self-consistency enter the estimator as inputs, in training and at inference alike, and the
estimator is never trained to imitate the guide (\cref{sec:dense}). Using flow as an input has
precedents. FlowNet2~\cite{ilg2017flownet2} stacks networks on the previous estimate and its warping
error, sparse external flow hints modulate the correlation of a recurrent
estimator~\cite{poggi2021sensorguided}, E-RAFT~\cite{gehrig2021eraft} warm-starts from the previous
interval, ResFlow~\cite{zhou2026resflow} refines a global motion estimate by residuals, an
event-based frame interpolator~\cite{chen2023revisiting} refines frame-based flow with events, and in
stereo a depth foundation model initializes the recurrent disparity~\cite{jiang2025defomstereo}. We show
that, for a guide from another sensor, its entry point decides the outcome (\cref{sec:fusionrules,sec:guidanceabl}).

\section{\ours{}}
\label{sec:method}
We estimate dense flow $\hat f\in\mathbb{R}^{2\times H\times W}$ over a
target interval $[t_0, t_0{+}\Delta t)$ from
two event windows, optionally guided by the frame pair spanning the same
interval. \cref{fig:arch} gives an overview.

\subsection{Event Estimator with a Depth Foundation Prior}
\label{sec:estimator}
\paragraph{Representation.} An event camera outputs a stream of events. The $m$-th
event $e_m=(x_m,y_m,t_m,p_m)$ is triggered at time $t_m$ when the log brightness at pixel
$(x_m,y_m)$ changes by a fixed contrast threshold, and its polarity $p_m\in\{-1,+1\}$
gives the sign of the change. Following E-RAFT~\cite{gehrig2021eraft}, we split the stream
into two windows, $\mathcal{E}_0$ in $[t_0{-}\Delta t,t_0)$ and $\mathcal{E}_1$ in
$[t_0,t_0{+}\Delta t)$, and convert each window $\mathcal{E}_k$ into a voxel grid
$V_k\in\mathbb{R}^{B\times H\times W}$, which divides the window into $B$ time bins of
$H\times W$ pixels. The cell at time bin $b\in\{0,\dots,B{-}1\}$ and pixel $(x,y)$ sums the
polarities of the nearby events $e_m\in\mathcal{E}_k$,
\begin{equation}
V_k(b,x,y)=\sum_m p_m\,\kappa(x{-}x_m)\,\kappa(y{-}y_m)\,\kappa(b{-}t_m^{*}),
\label{eq:voxel}
\end{equation}
where $t_m^{*}$ is $t_m$ scaled linearly to $[0,B{-}1]$ over the window, and
$\kappa(a)=\max(0,1{-}|a|)$ is a triangular weight that decreases linearly with the
distance between the event and the cell and vanishes beyond one cell. Finally, the
non-zero cells of each grid are normalized to zero mean and unit variance.

\paragraph{Correlation and Depth Features.} A shared ResNet-18 encoder maps each grid $V_k$
to a feature map $F_k$ at $1/8$ resolution, and a context encoder maps $[V_0,V_1]$ to the
initial hidden state $h^0$ and context $c$. To inject the depth prior, we render each grid as a
polarity image $I_k$ for the \emph{frozen} Depth AnyEvent
model~$\mathcal{D}$~\cite{bartolomei2025dae}, which returns decoder features and an inverse
depth map, $(\Phi_k,d_k)=\mathcal{D}(I_k)$. A trained head projects $\Phi_k$ to
$\bar\Phi_k$ on the grid of $F_k$, giving $\tilde F_k=[F_k,\bar\Phi_k]$. The correlation
pyramid compares every cell $\mathbf{x}_1$ of the first window with every cell
$\mathbf{x}_2$ of the second,
\begin{equation}
C_l(\mathbf{x}_1,\mathbf{x}_2)=\tfrac{1}{\sqrt{D'}}\big\langle \tilde F_0(\mathbf{x}_1),\,\tilde F_1^{(l)}(\mathbf{x}_2)\big\rangle,
\label{eq:corr}
\end{equation}
where $D'$ is the number of channels and $\tilde F_1^{(l)}$ is $\tilde F_1$ average-pooled by
$2^l$, giving four levels $l=0,\dots,3$. The lookup $L(C,f^k)$ samples every level within
radius $4$ around the match predicted by the current flow $f^k$. Following FlowSeek~\cite{poggi2025flowseek},
$d_0$ also defines eight motion bases $\mathcal{B}(d_0)$ that approximately span the flow of a
static scene for small camera motion: three translations scaled by $2d_0$ and five
depth-independent rotations, each of unit $\ell_2$ norm. A trained encoder maps
them to the hidden state and a basis context $c_b$.

\paragraph{Iterative Refinement.} Starting from $f^0$ (\cref{sec:dense}), the ConvNeXt
update operator $U$ and flow head $\Delta f$ run for $K$ iterations ($K{=}6$ in
training, $4$ at inference, \cref{sec:tta}),
\begin{equation}
\begin{split}
h^{k+1}&=U\big(h^k,\,[c,c_b],\,L(C,f^k),\,f^k\big),\\
f^{k+1}&=f^k+\Delta f(h^{k+1}).
\end{split}
\label{eq:update}
\end{equation}
Convex $8{\times}$ upsampling restores full resolution: each pixel is a convex combination of
the flow at its $3{\times}3$ neighboring cells, with softmax weights predicted from $h^k$.

\subsection{Frame Guide and Registration}
\label{sec:guide}
\paragraph{Guide Flow.} A \emph{frozen} frame-based FlowSeek-T~\cite{poggi2025flowseek}, built on Depth Anything
v2~\cite{yang2024dav2} and trained on synthetic data, Sintel, KITTI and HD1K but never on DSEC, runs
on the two RGB frames that bound the target interval, which DSEC time-aligns with the flow
ground truth ($12$ iterations, averaged with the mirrored pair). It gives the forward flow
$f^{\mathrm{F}}=f^{\mathrm{F}\rightarrow}$ from $t_0$ to $t_1=t_0{+}\Delta t$ and the
backward flow $f^{\mathrm{F}\leftarrow}$, at frame pixels $\mathbf{u}$.

\paragraph{Forward--Backward Gate.} Where the frame flow is correct, following it forward and
back returns to the start. The round-trip residual $r$ flags guide failures and occlusions,
and the gate $G$ keeps the pixels with $r<\tau{=}1.5$\,px,
\begin{equation}
\begin{split}
r(\mathbf{u})&=\big\|f^{\mathrm{F}\rightarrow}(\mathbf{u})
+f^{\mathrm{F}\leftarrow}\big(\mathbf{u}+f^{\mathrm{F}\rightarrow}(\mathbf{u})\big)\big\|,\\
G&=\mathbb{1}[\,r<\tau\,].
\end{split}
\label{eq:gate}
\end{equation}
This consistency check uses neither ground truth nor events, so it is available in training
and at test time.

\paragraph{Registration.} We warp the guide into the event view with the rotation-only
homography $\mathbf{H}=\mathbf{K}_e\mathbf{R}\mathbf{K}_f^{-1}$ ($\mathbf{K}_e,\mathbf{K}_f$
the intrinsics of the rectified event and frame cameras, $\mathbf{R}$ their relative
rotation), applied in homogeneous coordinates to both endpoints of each flow vector. This gives the guide
flow $f^{\mathrm{T}}$ in the event view,
\begin{equation}
f^{\mathrm{T}}(\mathbf{H}\mathbf{u})=\mathbf{H}\big(\mathbf{u}+f^{\mathrm{F}}(\mathbf{u})\big)-\mathbf{H}\mathbf{u},
\label{eq:warp}
\end{equation}
and resample $r$ and $G$ likewise. \cref{eq:warp} is exact for a pure rotation, but the camera
centers differ by $\mathbf{b}=(4.37,-0.05,1.36)$\,cm, and the error this causes has a closed
form.

\begin{proposition}[Registration Parallax]
\label{prop:parallax}
Assume rectified pinhole cameras with a common focal length $\varphi$ (in pixels) on a rigid,
synchronized rig. For a scene point with inverse depth $\rho$ and normalized image
coordinates $(\bar x,\bar y)$, let $\bdelta_t$ be the offset from its event-view projection
to its warped frame projection at time $t$. To first order in $\rho\|\mathbf{b}\|$,
$\bdelta=-\varphi\rho\,(b_x-\bar x b_z,\,b_y-\bar y b_z)$. For an exact frame flow
$f^{\mathrm{F}}$, at every event pixel $\mathbf{v}$ whose frame-visible point the event
camera also sees,
\begin{equation}
f^{\mathrm{T}}(\mathbf{v})=f(\mathbf{v}-\bdelta_0)
+(\bdelta_1-\bdelta_0),
\label{eq:parallax}
\end{equation}
where $f$ is the true flow in the event view.
\end{proposition}
\noindent The guide is thus the true flow shifted by $\bdelta_0$ ($|\bdelta|\approx25\rho$\,px,
$\rho$ in m$^{-1}$), plus a term $\bdelta_1{-}\bdelta_0$ that varies slowly over the image. At
a motion boundary, however, the discontinuity moves by the near side's $|\bdelta_0|$, leaving
an error band on \emph{one} side of every boundary, the left on DSEC. On validation, the guide
error within $4$\,px of vertical boundaries is $2.36$\,px on their left against $1.04$ on
their right, whereas the guided estimate is symmetric ($1.15$ against $1.17$). Computed in the
frame camera, where $\mathbf{b}$ plays no role, the gate still accepts $63\%$ of this band.

\paragraph{Guide Inputs During Training.} The same frozen guide
provides the guidance inputs of \cref{sec:dense} for every training
interval, computed offline in both directions together with the gate. It
is never a target (\cref{sec:training}). LiDAR ground truth covers $18\%$
of a training frame on average, whereas the gate accepts the guide on
$66\%$, so the estimator sees a dense, self-certified estimate on most
unsupervised pixels.

\subsection{Output-Level Fusion Baseline}
\label{sec:fusion}
Before feeding the guide into the estimator, we test the simplest alternative: blending the
two flows at the output. Let $\hat f$ be the prediction of the event-only estimator of
\cref{sec:estimator}. Wherever the gate accepts the guide, we mix the two flows with a single
weight $w\in[0,1]$, and elsewhere we keep $\hat f$,
\begin{equation}
f^{\star}=G\big[w\,f^{\mathrm{T}}+(1{-}w)\hat f\,\big]+(1{-}G)\,\hat f .
\label{eq:fusion}
\end{equation}
Blending beats replacing. Let $\sigma_g$ and $\sigma_e$ be the standard deviations of the
guide and event-only errors and $\eta$ their correlation. The error variance of the blend,
$w^2\sigma_g^2+(1{-}w)^2\sigma_e^2+2w(1{-}w)\eta\sigma_g\sigma_e$, is minimized at some
$w^{*}<1$ whenever $\eta<\sigma_g/\sigma_e$. Substituting the guide ($w{=}1$) therefore loses
to blending, even though the guide is the stronger source (\cref{sec:fusionrules}). We tuned
every component of this rule on the benchmark server: the guide resolution, the weight (flat
optimum at $w\in[0.65,0.72]$), and mirror and multi-scale averaging. Together they lower the
EPE from $0.759$ to $0.670$ (\cref{tab:stack}). The limitation is structural. On
gate-rejected pixels \cref{eq:fusion} falls back to $\hat f$, and a single $w$ cannot express
that the guide deserves more trust in the far field than on near structure
(\cref{sec:fusionrules}).

\subsection{Dense Guidance}
\label{sec:dense}
The blend uses the guide only after the estimator has finished. We instead give the guide to
the estimator as an input, so that it can decide per pixel how much to trust it. The guide
enters at two points, the context and the initial flow.

\paragraph{Guidance Input.} On the $1/8$ grid of the estimator, we stack the guide flow, the
gate and the residual,
\begin{equation}
\mathbf{g}=\big[\tfrac{1}{8}\mathcal{P}(f^{\mathrm{T}}),\;\mathcal{P}(G),\;
\tanh\!\big(\tfrac{1}{3}\mathcal{P}(r)\big)\big],
\label{eq:guide}
\end{equation}
where $\mathcal{P}$ is $8{\times}8$ average pooling, the factor $\frac{1}{8}$ converts the
guide flow to grid units, and $\tanh$ bounds the residual $r$ of \cref{eq:gate}. A small
convolutional guidance encoder $E$ adds $\mathbf{g}$ to the context of \cref{eq:update},
$[c,c_b]\leftarrow[c,c_b]+E(\mathbf{g})$, so every update step sees the guide, its gate and
its residual.

\paragraph{Guided Initialization.} Without guidance, the estimator starts from its own guess
$f^{\text{init}}$, the flow head applied to $h^0$. We start it from the guide wherever the gate
accepts,
\begin{equation}
f^{0}=\mathcal{P}(G)\cdot\tfrac{1}{8}\mathcal{P}(f^{\mathrm{T}})
+\big(1-\mathcal{P}(G)\big)\cdot f^{\text{init}}.
\label{eq:guideinit}
\end{equation}
Three properties follow. \emph{(i) Rejected pixels keep the estimator's own start}, while $r$
stays visible to the update operator as evidence. \emph{(ii) The start, not the search range,
binds.} The trained update recovers initial errors only up to about $16$\,px
(\cref{sec:basin}), so a fast pixel must start near its match. \emph{(iii) Training starts from
the unguided model.} The last layer of $E$ is zero-initialized, so a guided model restored from
an unguided checkpoint differs from it only through the start of \cref{eq:guideinit}.

Dense guidance \emph{replaces} the blend rather than supplementing it: a
guided model scores $0.636$ on the test server against the blend's $0.670$,
and the blend applied on top regresses it to $0.662$. The cost is that guidance ties the weights
to the guide: without it the guided model is worse than the event-only one,
whereas the blend falls back to $\hat f$.

\subsection{Training}
\label{sec:training}
\paragraph{Supervised Loss.} The estimator outputs $K{+}1$ predictions $f^k$, from the initial
flow $f^0$ to the last update, each scored by a likelihood as in SEA-RAFT~\cite{wang2024searaft}. The flow head also predicts, per pixel, mixture
logits $\alpha\in\mathbb{R}^2$ and a log-scale $s_1$, and the residual $e$ between prediction
and ground truth is modeled per axis as a two-component Laplace mixture,
$p(e)=\sum_j\tilde\alpha_j\,(2\beta_j)^{-1}\exp(-|e|/\beta_j)$, with
$\tilde\alpha=\operatorname{softmax}(\alpha)$ and $\beta_j=\exp(s_j)$. One component models
outliers ($s_1\in[0,10]$), the other inliers ($s_2=0$). With $\ell^k$ the negative log-likelihood of $f^k$ and $M$ the ground-truth validity mask, a
weight $\omega=1+\lambda_e\,q/\max q$ grows with the ground-truth flow gradient $q$ at motion
boundaries.

\paragraph{Self-Supervised Loss.} Three self-supervised terms cover the
unlabeled pixels. Contrast maximization ($\mathcal{L}_{\text{cm}}$) warps the
events by the predicted flow and rewards sharp event counts. Edge-aware smoothness
($\mathcal{L}_{\text{sm}}$) and a penalty on local contraction,
$\mathcal{L}_{\text{div}}=\max(0,-\nabla\!\cdot\!\hat f)^2$, regularize the flow, the latter
preventing the known collapse of contrast maximization. With
$\mathcal{L}_{\text{ss}}=\mathcal{L}_{\text{cm}}+\lambda_{\text{sm}}\mathcal{L}_{\text{sm}}+\lambda_{\text{div}}\mathcal{L}_{\text{div}}$, the objective is
\begin{equation}
\mathcal{L}=\sum_{k=0}^{K}\gamma^{K-k}\,\frac{\sum_{\mathbf{x}}\omega M\ell^k}{\sum_{\mathbf{x}}\omega M}+\lambda_{\text{ss}}\,\mathcal{L}_{\text{ss}},
\label{eq:total}
\end{equation}
with $\gamma{=}0.85$, $\lambda_e{=}1$, $\lambda_{\text{ss}}{=}5$,
$\lambda_{\text{sm}}{=}0.05$ and $\lambda_{\text{div}}{=}0.1$.

\paragraph{Schedule.} The estimator is initialized from the frame-based FlowSeek-T weights and
trained in two stages: $60$k steps on a streamed subset of MultiFlow~\cite{gehrig2024dense},
then $120$k steps on DSEC with guidance. We use batch size $6$, $320{\times}448$ crops, AdamW
with a OneCycle peak learning rate of $2{\times}10^{-4}$ and bfloat16 mixed precision on a
single GPU. The checkpoint with the lowest validation EPE, at step $115$k, is the
\emph{single model}. It produces
the server rows of \cref{tab:stack} from $0.607$ to $0.583$ and every guided DSEC validation
number, and its step-$35$k checkpoint produces the rows at $0.636$ and $0.612$ and the blend on
top ($0.662$). \ours{} (E), the event-only model, is trained for $120$k steps on DSEC alone, without
guidance or the MultiFlow stage. All ablations are separate runs on matched schedules.

\paragraph{Five-Model Ensemble.} The full system averages five models that are also trained on
large motion. About half of their samples span $n{\times}100$\,ms ($n\in\{2,3\}$), with chained
ground truth and the guide run over the whole span. Three
models are trained from the MultiFlow stage and two are fine-tuned from the single model.
Since DSEC frames arrive at $20$\,Hz, a frame lies at the middle of every interval, and the
models use three sets of guide inputs that read only frames inside the interval: (a)
FlowSeek-T on the two end frames, (b) a per-pixel forward--backward selection between (a) and
FlowSeek-T chained through the middle frame, and (c) MEMFOF~\cite{bargatin2025memfof} on all
three frames, with one quarter of the weight. Averaging reduces the run-to-run spread of single
models on the fastest sequence. A sixth run raised the server error in
every ensemble it joined and was left out.

\subsection{Symmetric Test-Time Ensembling}
\label{sec:tta}
Three parameter-free steps refine every test prediction (\cref{tab:stack}).
\emph{(i) Symmetric ensemble.} We average the predictions on the original input, its horizontal
mirror image, and both at $0.75\times$ scale, each mapped back to the original frame, with the
guide inputs transformed together with the events. Downscaling brings large displacements into
the range the refinement can recover (\cref{sec:basin}). \emph{(ii) Fewer iterations.} We run
four update iterations instead of the six used in training. \emph{(iii) Epipolar projection.}
As DSEC ground truth is the flow of a static scene under ego-motion, we project every flow
vector onto its epipolar line under an essential matrix fitted to the prediction.

\section{Experiments}
\label{sec:experiments}
\paragraph{Data and Protocol.} DSEC-Flow~\cite{gehrig2021dsec,gehrig2021eraft} is the standard
benchmark for event-based optical flow. It provides $8{,}170$ forward-flow pairs over $18$
driving sequences ($480{\times}640$, $100$\,ms) with LiDAR ground truth and validity masks. We
train on $16$ sequences ($6{,}722$ pairs) and hold out \texttt{zurich\_city\_05\_a} and
\texttt{zurich\_city\_11\_c} ($1{,}448$ pairs) for validation. The $416$ test frames
($7$ sequences, including night and high-speed rural driving) have hidden ground truth and are
scored by the benchmark server: endpoint error (EPE), $N$\,px outlier rate ($N$PE) and angular
error (AE). MVSEC~\cite{zhu2018mvsec} follows the standard protocol (\cref{sec:mvsec}).

\begin{table}[t]
\centering
\small
\setlength{\tabcolsep}{4.2pt}
\resizebox{\ifdim\width>\linewidth\linewidth\else\width\fi}{!}{%
\begin{tabular}{llccccc}
\toprule
Method & Input & EPE$\downarrow$ & 1PE$\downarrow$ & 2PE$\downarrow$ & 3PE$\downarrow$ & AE$\downarrow$ \\
\midrule
RAFT~\cite{teed2020raft} (frame-only)$^{\ddagger}$ & I & $0.78$ & $12.40$ & $4.60$ & $2.61$ & $2.44$ \\
\midrule
E-RAFT~\cite{gehrig2021eraft} & E & $0.788$ & $12.74$ & $4.74$ & $2.68$ & $2.85$ \\
\ours{} (E) & E & $0.759$ & $11.20$ & $4.02$ & $2.44$ & $2.60$ \\
TMA~\cite{liu2023tma} & E & $0.743$ & $10.86$ & $3.97$ & $2.30$ & $2.68$ \\
EDCFlow~\cite{liu2025edcflow} & E & $0.72$ & $10.0$ & $3.6$ & $2.1$ & $2.65$ \\
IDNet~\cite{wu2024idnet} & E & $0.719$ & $10.07$ & $3.50$ & $2.04$ & $2.72$ \\
ECDDP$^{\dagger}$~\cite{yang2024ecddp} & E & $0.697$ & $8.89$ & $3.20$ & $1.96$ & $2.58$ \\
BFlow~\cite{gehrig2024dense} & E+I & $0.69$ & $9.70$ & $3.42$ & $1.88$ & $2.42$ \\
P-SSE~\cite{psse2025} & E & $0.680$ & $9.14$ & $3.23$ & $1.82$ & $2.56$ \\
STSC-Flow~\cite{hu2026stsc} & E & $0.663$ & $7.94$ & $2.67$ & $1.60$ & $2.53$ \\
STFlow~\cite{stflow2025} & E+I & $0.66$ & $8.58$ & $2.93$ & $1.68$ & $2.37$ \\
BAT~\cite{xu2026bat} & E & $0.650$ & $7.54$ & $2.84$ & $1.74$ & $2.43$ \\
EMatch~\cite{zhang2025ematch} & E & $0.640$ & $7.86$ & $2.83$ & $1.69$ & $2.46$ \\
STFlow$^{\dagger}$~\cite{stflow2025} & E+I & $0.630$ & $7.93$ & $2.61$ & $\underline{1.45}$ & $2.29$ \\
EDCPT$^{\dagger}$~\cite{wan2025edcpt} & E+I & $0.625$ & $\underline{6.87}$ & $\underline{2.35}$ & $1.52$ & $\underline{2.17}$ \\
E-WAVE$^{\dagger}$~\cite{wu2026ewave} & E+I & $\underline{0.623}$ & $7.26$ & $2.54$ & $1.56$ & $2.24$ \\
\midrule
\textbf{\ours}$^{\dagger}$ (E+I), single model & E+I & $0.583$ & $6.31$ & $2.06$ & $1.18$ & $2.21$ \\
\textbf{\ours}$^{\dagger}$ (E+I), 5-model ensemble & E+I & $\mathbf{0.554}$ & $\mathbf{6.01}$ & $\mathbf{1.94}$ & $\mathbf{1.11}$ & $\mathbf{2.08}$ \\
\bottomrule
\end{tabular}}
\caption{\textbf{DSEC-Flow Test Server} ($416$ frames in $7$ sequences, hidden ground truth),
sorted by decreasing EPE. Bold marks the best result, underline the best prior method. Input: E
events, I RGB frames. $^{\dagger}$Pretrained on synthetic event data. $^{\ddagger}$Frame-only, as
reported by BFlow~\cite{gehrig2024dense}. Both \ours{} rows use the test-time steps of
\cref{sec:tta}, with the guide and depth models frozen and pretrained on other data.}
\label{tab:leaderboard}
\end{table}

\begin{table}[t]
\centering
\small
\setlength{\tabcolsep}{5pt}
\resizebox{\ifdim\width>\linewidth\linewidth\else\width\fi}{!}{%
\begin{tabular}{lcc}
\toprule
Configuration (test server) & EPE$\downarrow$ & $\Delta$ \\
\midrule
\ours{} (E), \cref{sec:estimator}                     & $0.759$ & -- \\
$+$ gated output blend, $w{=}0.85$, $2/3$-res.\ guide (\cref{eq:fusion}) & $0.706$ & $-7.0\%$ \\
$+$ full-resolution guide, $12$ instead of $4$ guide iterations & $0.689$ & $-2.4\%$ \\
$+$ $w$ at the server-measured plateau ($0.70$)                 & $0.681$ & $-1.2\%$ \\
$+$ mirror averaging, both branches                             & $0.678$ & $-0.4\%$ \\
$+$ multi-scale averaging, event branch                         & $0.670$ & $-1.2\%$ \\
\midrule
Dense guidance replaces the blend, step $35$k (\cref{sec:dense}) & $0.636$ & $-5.1\%$ \\
$+$ symmetric ensemble, four members (\cref{sec:tta})           & $0.612$ & $-3.8\%$ \\
$+$ single model, step-$115$k checkpoint (\cref{sec:training}) & $0.607$ & $-0.8\%$ \\
$+$ four update iterations at inference (\cref{sec:tta})        & $0.587$ & $-3.3\%$ \\
$+$ epipolar projection (\cref{sec:tta})                        & $0.583$ & $-0.7\%$ \\
$+$ average of two large-motion models (\cref{sec:training}) & $0.564$ & $-3.3\%$ \\
$+$ 5-model ensemble, three guide-input sets (\cref{sec:training}) & $\mathbf{0.554}$ & $-1.8\%$ \\
\bottomrule
\end{tabular}}
\caption{\textbf{Ablation on the DSEC Test Set}, from \ours{} (E) to our final E+I method.
Each row adds one change and is a separate server evaluation. The upper block builds the
strongest output-level fusion we could find (\cref{sec:fusion}), and the lower block replaces it
with dense guidance. Dense guidance alone ($0.759\to0.636$) gains more than the upper block
($0.759\to0.670$), and the gated blend on top of the guided model \emph{degrades} it
($0.636\to0.662$, $+4.1\%$).}
\label{tab:stack}
\end{table}

\begin{figure*}[t]
\centering
\includegraphics[width=\textwidth]{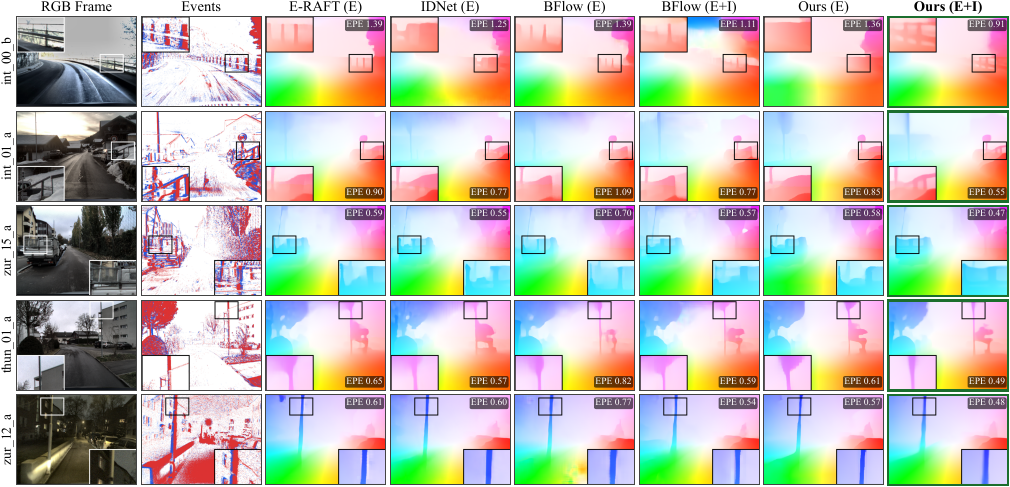}
\caption{\textbf{Qualitative Comparison on the DSEC Test Sequences}, day and night. Each box is enlarged in the inset of the same tile. Since the test ground truth is hidden, badges give each method's benchmark EPE on that sequence. The guard rail, the railing, the truck and the poles keep their shape in \ours{} (E+I), while the other methods spread them into their surroundings. Baselines are run from their publicly available checkpoints. Ours (E) is the event-only model and Ours (E+I) our final model. Inputs: (E) events, (E+I) events and RGB frames. Please zoom in for details.}
\label{fig:qual_main}
\end{figure*}

\subsection{DSEC-Flow Benchmark}
\label{sec:benchmark}
\ours{} ranks first of $39$ entries by EPE (\cref{tab:leaderboard}, November 9, 2026), with
$0.554$ EPE, $6.01\%$ $1$PE, $1.94\%$ $2$PE, $1.11\%$ $3$PE and $2.08^\circ$ AE. Even without model averaging, the single model of \cref{sec:training} with the test-time
steps of \cref{sec:tta} scores $0.583$ and leads every published method on all metrics except
AE, where EDCPT~\cite{wan2025edcpt} has $2.17^\circ$ against $2.21^\circ$. Per sequence,
\ours{} has the lowest or tied-lowest EPE of the published methods on six of seven sequences
(EDCPT is lower on \texttt{thun\_01\_b}).
\cref{fig:qual_main} shows the difference at the pixel level, and \cref{fig:dsecscatter} places
every published method by EPE and 3PE.

\cref{tab:stack} decomposes the result into server evaluations.

\subsection{Output-Level Fusion Versus Dense Guidance}
\label{sec:fusionrules}
Is dense guidance better than the best output-level fusion, and why? We compare every fusion
rule against ground truth on the full validation split. \emph{First}, the guide beats the
event-only estimate on its gated pixels ($0.484$ against $0.510$\,px), yet substituting it
($0.520$) loses to blending ($0.487$), as \cref{sec:fusion} predicts. \emph{Second}, the best
output rule stays far from a per-pixel oracle that picks the better source at every pixel
($0.487$ against $0.388$): one weight serves every pixel, and on gate-rejected pixels every
output rule returns the event estimate unchanged. \emph{Third}, dense guidance ($0.435$), with
no parameter fitted at inference, closes $53\%$ of this gap, and it does so where both sources
are available. On gate-accepted pixels it reaches $0.394$ against $0.450$ for the best blend,
while on the $3\%$ of labeled pixels the gate rejects, it is $5\%$ worse than the event-only
model ($1.80$ against $1.71$\,px). The frames alone do not explain the result: RAFT, FlowSeek-T
and MEMFOF, run on frames only with the same registration, score $0.527$--$0.538$ on this split,
close to the event-only model ($0.546$).

The two sources are complementary: the guide's margin over the event-only model shrinks from
$23\%$ in the far field to zero at the near end, where the parallax band of
\cref{prop:parallax} is widest, and the guided estimate matches the guide in the far field and
is below both sources elsewhere.

\begin{table}[t]
\centering
\small
\setlength{\tabcolsep}{4pt}
\resizebox{\ifdim\width>\linewidth\linewidth\else\width\fi}{!}{%
\begin{tabular}{lcccccc}
\toprule
Guidance inputs & $f^{\mathrm{T}}$ & $G$ & $r$ & Init & Best EPE$\downarrow$ & Final EPE$\downarrow$ \\
\midrule
None (events only) & -- & -- & -- & -- & $0.569$ & $0.574$ \\
Guide flow only & $\checkmark$ & -- & -- & -- & $0.468$ & $0.474$ \\
$+$ gate, no residual & $\checkmark$ & $\checkmark$ & -- & $\checkmark$ & $0.434$ & $0.439$ \\
$+$ residual, no gate & $\checkmark$ & -- & $\checkmark$ & $\checkmark$ & $0.437$ & $0.439$ \\
Full, no guide init & $\checkmark$ & $\checkmark$ & $\checkmark$ & -- & $0.467$ & $0.474$ \\
Full (\cref{eq:guide,eq:guideinit}) & $\checkmark$ & $\checkmark$ & $\checkmark$ & $\checkmark$ & $0.434$ & $0.437$ \\
\bottomrule
\end{tabular}}
\caption{\textbf{Guidance-Input Ablation.} Held-out validation (every fourth
frame), matched 40k schedule from the MultiFlow stage. Disabled inputs are zeroed.
$f^{\mathrm{T}}$, $G$, $r$: warped guide flow, gate and residual as context
inputs. Init: flow state starts from the guide where the gate accepts
(\cref{eq:guideinit}), even without $G$. ``Final'': mean of the last three validations.
Separate $40$k-step runs, so absolute values differ from the deployed models of
\cref{sec:fusionrules}.}
\label{tab:guidance}
\end{table}

\subsection{Guidance-Input Ablation}
\label{sec:guidanceabl}
Which guidance inputs matter? \cref{tab:guidance} removes the inputs of
\cref{eq:guide,eq:guideinit} one at a time, feeding zeros instead, on a matched $40$k-step
schedule from the MultiFlow stage. The warped guide flow alone, entering only through the
context, closes most of the gap to the full model ($0.569\to0.468$). Adding the gate and the
residual to the context without the initialization changes nothing ($0.467$). The
initialization of \cref{eq:guideinit} accounts for the rest ($0.434$--$0.437$), and once the
flow starts from the guide, either trust signal suffices ($0.434$ against $0.437$). Each row is
a single run whose last three validations vary by at most $0.02$\,px, so the initialization
effect exceeds the fluctuation within a run (seeds were not repeated).

\begin{figure*}[t]
\centering
\includegraphics[width=\textwidth]{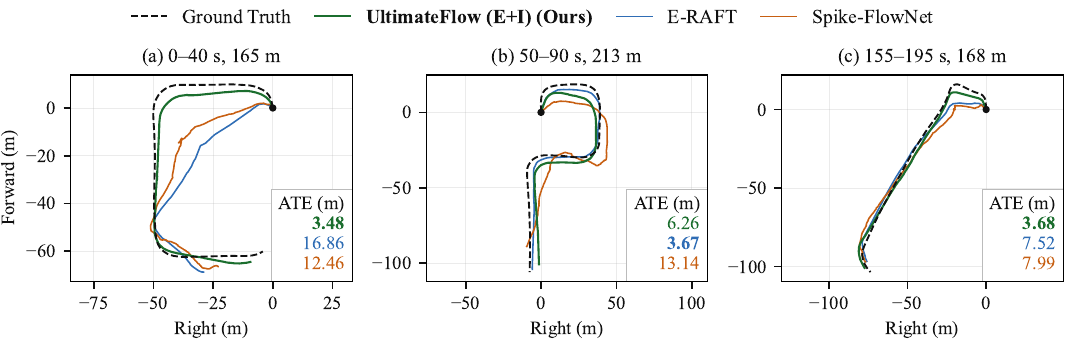}
\caption{\textbf{Trajectories from Flow} on MVSEC \texttt{od1}: the three non-overlapping $40$\,s windows with the largest heading change, each aligned by one similarity transform. ATE of each method in its color, best in bold.}
\label{fig:vo}
\end{figure*}

\subsection{Convergence Basin}
\label{sec:basin}
How far can the update move a wrong start? We start each model at the ground truth plus a
global offset of $\epsilon$\,px and record the error after each update. The trained event-only
estimator removes offsets up to $\epsilon{=}16$\,px within six updates ($0.80$ against
$0.51$\,px from an exact start) but not larger ones, although its coarsest correlation level
spans $\pm256$\,px. Its refinement is local, so the start bounds the result on large motion. The
guided model is more local still: it refines its guide start ($0.510\to0.414$\,px) and corrects
errors confined to a band at a motion boundary, such as the parallax of \cref{prop:parallax},
but not an offset of the whole image. On real large motion, $300$\,ms windows of the held-out
sequences, the guided model reaches $0.842$\,px against $1.174$ for the event-only model, and
switching off its initialization at inference raises it to $2.254$.

\paragraph{Depth Prior.}
\label{sec:depthabl}
Retraining an earlier event-only recipe with the prior's inputs zeroed shows that the prior
lowers the final EPE by $8.3\%$ ($0.697\to0.639$), while removing only the rigid-motion bases
does not hurt ($0.635$). The gain thus comes from the prior's \emph{features} in the correlation
volume, not from the motion bases. The prior was fine-tuned on DSEC depth that includes
\texttt{zurich\_city\_11\_c}, so this validation gain may be optimistic.

\begin{table}[t]
\centering
\small
\setlength{\tabcolsep}{6pt}
\begin{tabular}{lcccc}
\toprule
Method & \texttt{if1} & \texttt{if2} & \texttt{if3} & \texttt{od1} \\
\midrule
EV-FlowNet~\cite{zhu2018evflownet} & $1.03$ & $1.72$ & $1.53$ & $0.49$ \\
E-RAFT~\cite{gehrig2021eraft}      & $1.10$ & $1.94$ & $1.66$ & $\underline{0.24}$ \\
TMA~\cite{liu2023tma}              & $1.06$ & $1.81$ & $1.58$ & $0.25$ \\
EDCFlow~\cite{liu2025edcflow}      & $\underline{0.97}$ & $\underline{1.49}$ & $\underline{1.38}$ & $\mathbf{0.23}$ \\
\textbf{\ours} (E+I)               & $\mathbf{0.84}$ & $\mathbf{1.47}$ & $\mathbf{1.27}$ & $0.27$ \\
\bottomrule
\end{tabular}
\caption{\textbf{MVSEC} over one frame interval, $dt{=}1$ (EPE on pixels with
ground truth and events, best in bold, second best underlined, \texttt{if} =
\texttt{indoor\_flying}, \texttt{od} = \texttt{outdoor\_day}). Training on
\texttt{od2}, testing on the rest. Baselines as published, from
the supervised methods tabulated by EDCFlow~\cite{liu2025edcflow}.
Self-supervised, model-based and event-plus-frame methods outside this table
report lower indoor EPE~\cite{zhu2019unsupervised,lee2020spikeflownet}.}
\label{tab:mvsec}
\end{table}

\subsection{Fine-Tuned Evaluation on MVSEC}
\label{sec:mvsec}
MVSEC's DAVIS sensor records frames and events on \emph{one} pixel array, so $\mathbf{b}{=}0$
and \cref{prop:parallax} predicts no parallax band. Its \texttt{if} sequences are
hexacopter flights, far from the driving domain of DSEC. Under the standard protocol, \ours{},
fine-tuned with frame guidance over one to three frame intervals, has the lowest EPE of the
methods in \cref{tab:mvsec} on all three \texttt{if} sequences. It trails the best EPE by
$0.04$\,px on \texttt{od1}.

\subsection{Camera Motion from Flow}
\label{sec:odometry}
To test flow on a downstream task, we fit an essential matrix to
each method's flow on every consecutive-frame interval of MVSEC \texttt{od1} and chain
the resulting poses. Under the KITTI protocol, \ours{} has the lowest translation drift ($20.5\%$ against $27.9\%$ for
E-RAFT~\cite{gehrig2021eraft} and $31.2\%$ for Spike-FlowNet~\cite{lee2020spikeflownet}), rotation drift
and $1$\,s relative pose error, and the lowest absolute trajectory error (ATE, after a
similarity alignment for the unknown scale) in $66\%$ of all $20$\,s windows (E-RAFT $30\%$, Spike-FlowNet $4\%$). \cref{fig:vo} shows three windows. Indoors, however,
\ours{} trails Spike-FlowNet on the translation direction.

\paragraph{Robustness and Cost.}
\label{sec:analysis}
Removing $90\%$ of the events raises the EPE of the guided model by $8\%$ and of the
event-only model by $66\%$, and a tenfold darker frame raises it by $6\%$. The failure mode is
frame noise that collapses the gate. The estimator has $18.2$M trained parameters and runs in
$38$\,ms per pair at $K{=}4$ on one RTX~5080.

\section{Conclusion and Limitations} \label{sec:conclusion}
We introduced \ours{}, which uses a frozen RGB optical flow network not as a second estimate to blend at the output, but as a dense input to an event-based estimator built on an event depth foundation model. The guide flow, its forward--backward residual and the gate enter the estimator, and the recurrent refinement starts from the guide wherever it is reliable. This outperforms every output-level fusion we could build, and \ours{} ranks first on the DSEC-Flow benchmark with an EPE of $0.554$. More broadly, a frozen model of another sensor is most useful as an input and a starting point for the estimator, not as a second opinion blended at its output.

The guided model depends on its guide: without it, or when frame noise collapses the gate, it falls below the event-only model. The guide reads the frame at the end of the interval and dominates the runtime, so the method is neither causal nor low latency, and it needs calibrated extrinsics. A causal guide that reads only past frames, and a lighter one, are the natural next steps. As frame-based foundation models keep improving, event cameras can now inherit that progress without new labels, and see in the fast and dark scenes where frames alone fail. Code and checkpoints will be released.

\section*{Acknowledgments}
This work was supported by a Google Cloud Research Credits award from Google.

\makeatletter\typeout{REFS-START-PAGE \thepage\space col \if@firstcolumn left\else right\fi\space at \the\pagetotal\space of \the\textheight}\makeatother
{
    \small
    \bibliographystyle{ieeenat_fullname}
    \bibliography{references}

\begin{thebibliography}{38}
\providecommand{\natexlab}[1]{#1}
\providecommand{\url}[1]{\texttt{#1}}
\expandafter\ifx\csname urlstyle\endcsname\relax
  \providecommand{\doi}[1]{doi: #1}\else
  \providecommand{\doi}{doi: \begingroup \urlstyle{rm}\Url}\fi

\bibitem[Bargatin et~al.(2025)Bargatin, Chistov, Yakovenko, and
  Vatolin]{bargatin2025memfof}
Vladislav Bargatin, Egor Chistov, Alexander Yakovenko, and Dmitriy Vatolin.
\newblock {MEMFOF}: High-resolution training for memory-efficient multi-frame
  optical flow estimation.
\newblock In \emph{Proceedings of the IEEE/CVF International Conference on
  Computer Vision (ICCV)}, 2025.

\bibitem[Bartolomei et~al.(2025)Bartolomei, Mannocci, Tosi, Poggi, and
  Mattoccia]{bartolomei2025dae}
Luca Bartolomei, Enrico Mannocci, Fabio Tosi, Matteo Poggi, and Stefano
  Mattoccia.
\newblock {Depth AnyEvent}: A cross-modal distillation paradigm for event-based
  monocular depth estimation.
\newblock In \emph{Proceedings of the IEEE/CVF International Conference on
  Computer Vision (ICCV)}, 2025.

\bibitem[Bochkovskii et~al.(2025)Bochkovskii, Delaunoy, Germain, Santos, Zhou,
  Richter, and Koltun]{bochkovskii2025depthpro}
Aleksei Bochkovskii, Ama{\"e}l Delaunoy, Hugo Germain, Marcel Santos, Yichao
  Zhou, Stephan~R. Richter, and Vladlen Koltun.
\newblock {Depth Pro}: Sharp monocular metric depth in less than a second.
\newblock In \emph{International Conference on Learning Representations
  (ICLR)}, 2025.

\bibitem[Chen et~al.(2023)Chen, Zhu, Lian, Yang, Wang, Zhang, Liu, Qian, Kneip,
  and Gao]{chen2023revisiting}
Jiaben Chen, Yichen Zhu, Dongze Lian, Jiaqi Yang, Yifu Wang, Renrui Zhang,
  Xinhang Liu, Shenhan Qian, Laurent Kneip, and Shenghua Gao.
\newblock Revisiting event-based video frame interpolation.
\newblock In \emph{IEEE/RSJ International Conference on Intelligent Robots and
  Systems (IROS)}, 2023.

\bibitem[Gallego et~al.(2022)Gallego, Delbr{\"u}ck, Orchard, Bartolozzi, Taba,
  Censi, Leutenegger, Davison, Conradt, Daniilidis, and
  Scaramuzza]{gallego2022survey}
Guillermo Gallego, Tobi Delbr{\"u}ck, Garrick Orchard, Chiara Bartolozzi, Brian
  Taba, Andrea Censi, Stefan Leutenegger, Andrew~J. Davison, J{\"o}rg Conradt,
  Kostas Daniilidis, and Davide Scaramuzza.
\newblock Event-based vision: A survey.
\newblock \emph{IEEE Transactions on Pattern Analysis and Machine
  Intelligence}, 44\penalty0 (1):\penalty0 154--180, 2022.

\bibitem[Gehrig et~al.(2021{\natexlab{a}})Gehrig, Aarents, Gehrig, and
  Scaramuzza]{gehrig2021dsec}
Mathias Gehrig, Willem Aarents, Daniel Gehrig, and Davide Scaramuzza.
\newblock {DSEC}: A stereo event camera dataset for driving scenarios.
\newblock \emph{IEEE Robotics and Automation Letters}, 2021{\natexlab{a}}.

\bibitem[Gehrig et~al.(2021{\natexlab{b}})Gehrig, Millh{\"a}usler, Gehrig, and
  Scaramuzza]{gehrig2021eraft}
Mathias Gehrig, Mario Millh{\"a}usler, Daniel Gehrig, and Davide Scaramuzza.
\newblock {E-RAFT}: Dense optical flow from event cameras.
\newblock In \emph{International Conference on 3D Vision (3DV)},
  2021{\natexlab{b}}.

\bibitem[Gehrig et~al.(2024)Gehrig, Muglikar, and Scaramuzza]{gehrig2024dense}
Mathias Gehrig, Manasi Muglikar, and Davide Scaramuzza.
\newblock Dense continuous-time optical flow from event cameras.
\newblock \emph{IEEE Transactions on Pattern Analysis and Machine
  Intelligence}, 2024.

\bibitem[Govinda~Raju et~al.(2025)Govinda~Raju, Zubi{\'c}, Cannici, and
  Scaramuzza]{psse2025}
Gokul~Raju Govinda~Raju, Nikola Zubi{\'c}, Marco Cannici, and Davide
  Scaramuzza.
\newblock Perturbed state space feature encoders for optical flow with event
  cameras.
\newblock In \emph{Proceedings of the IEEE/CVF Conference on Computer Vision
  and Pattern Recognition Workshops (CVPRW)}, 2025.

\bibitem[Hu et~al.(2026)Hu, Wu, Yang, and Wu]{hu2026stsc}
Rui Hu, Song Wu, Wen Yang, and Jinjian Wu.
\newblock From contrast to consistency: Rethinking event-based continuous-time
  optical flow estimation.
\newblock In \emph{Proceedings of the IEEE/CVF Conference on Computer Vision
  and Pattern Recognition (CVPR)}, 2026.

\bibitem[Ilg et~al.(2017)Ilg, Mayer, Saikia, Keuper, Dosovitskiy, and
  Brox]{ilg2017flownet2}
Eddy Ilg, Nikolaus Mayer, Tonmoy Saikia, Margret Keuper, Alexey Dosovitskiy,
  and Thomas Brox.
\newblock {FlowNet} 2.0: Evolution of optical flow estimation with deep
  networks.
\newblock In \emph{Proceedings of the IEEE Conference on Computer Vision and
  Pattern Recognition (CVPR)}, 2017.

\bibitem[Jiang et~al.(2025)Jiang, Lou, Ding, Xu, Tan, Jiang, and
  Huang]{jiang2025defomstereo}
Hualie Jiang, Zhiqiang Lou, Laiyan Ding, Rui Xu, Minglang Tan, Wenjie Jiang,
  and Rui Huang.
\newblock {DEFOM-Stereo}: Depth foundation model based stereo matching.
\newblock In \emph{Proceedings of the IEEE/CVF Conference on Computer Vision
  and Pattern Recognition (CVPR)}, pages 21857--21867, 2025.

\bibitem[Lee et~al.(2020)Lee, Kosta, Zhu, Chaney, Daniilidis, and
  Roy]{lee2020spikeflownet}
Chankyu Lee, Adarsh~Kumar Kosta, Alex~Zihao Zhu, Kenneth Chaney, Kostas
  Daniilidis, and Kaushik Roy.
\newblock {Spike-FlowNet}: Event-based optical flow estimation with
  energy-efficient hybrid neural networks.
\newblock In \emph{Proceedings of the European Conference on Computer Vision
  (ECCV)}, 2020.

\bibitem[Li et~al.(2023)Li, Huang, Chen, Shi, Li, Bao, Cui, and
  Zhang]{li2023blinkflow}
Yijin Li, Zhaoyang Huang, Shuo Chen, Xiaoyu Shi, Hongsheng Li, Hujun Bao,
  Zhaopeng Cui, and Guofeng Zhang.
\newblock {BlinkFlow}: A dataset to push the limits of event-based optical flow
  estimation.
\newblock In \emph{IEEE/RSJ International Conference on Intelligent Robots and
  Systems (IROS)}, 2023.

\bibitem[Liu et~al.(2025)Liu, Cheng, Wang, and Sun]{liu2025edcflow}
Daikun Liu, Lei Cheng, Teng Wang, and Changyin Sun.
\newblock {EDCFlow}: Exploring temporally dense difference maps for event-based
  optical flow estimation.
\newblock In \emph{Proceedings of the IEEE/CVF Conference on Computer Vision
  and Pattern Recognition (CVPR)}, 2025.

\bibitem[Liu et~al.(2023)Liu, Chen, Qu, Zhang, Li, Knoll, and
  Jiang]{liu2023tma}
Haotian Liu, Guang Chen, Sanqing Qu, Yanping Zhang, Zhijun Li, Alois Knoll, and
  Changjun Jiang.
\newblock {TMA}: Temporal motion aggregation for event-based optical flow.
\newblock In \emph{Proceedings of the IEEE/CVF International Conference on
  Computer Vision (ICCV)}, 2023.

\bibitem[Luo et~al.(2024)Luo, Luo, Wang, Lin, Zeng, and Liu]{luo2024eemflow}
Xinglong Luo, Ao Luo, Zhengning Wang, Chunyu Lin, Bing Zeng, and Shuaicheng
  Liu.
\newblock Efficient meshflow and optical flow estimation from event cameras.
\newblock In \emph{Proceedings of the IEEE/CVF Conference on Computer Vision
  and Pattern Recognition (CVPR)}, 2024.

\bibitem[Poggi and Tosi(2025)]{poggi2025flowseek}
Matteo Poggi and Fabio Tosi.
\newblock {FlowSeek}: Optical flow made easier with depth foundation models and
  motion bases.
\newblock In \emph{Proceedings of the IEEE/CVF International Conference on
  Computer Vision (ICCV)}, 2025.

\bibitem[Poggi et~al.(2021)Poggi, Aleotti, and
  Mattoccia]{poggi2021sensorguided}
Matteo Poggi, Filippo Aleotti, and Stefano Mattoccia.
\newblock Sensor-guided optical flow.
\newblock In \emph{Proceedings of the IEEE/CVF International Conference on
  Computer Vision (ICCV)}, pages 7888--7898, 2021.

\bibitem[Teed and Deng(2020)]{teed2020raft}
Zachary Teed and Jia Deng.
\newblock {RAFT}: Recurrent all-pairs field transforms for optical flow.
\newblock In \emph{Proceedings of the European Conference on Computer Vision
  (ECCV)}, 2020.

\bibitem[Wan et~al.(2022)Wan, Dai, and Mao]{wan2022dceiflow}
Zhexiong Wan, Yuchao Dai, and Yuxin Mao.
\newblock Learning dense and continuous optical flow from an event camera.
\newblock \emph{IEEE Transactions on Image Processing}, 31:\penalty0
  7237--7251, 2022.

\bibitem[Wan et~al.(2025)Wan, Luo, Dai, and Lee]{wan2025edcpt}
Zhexiong Wan, Jianqin Luo, Yuchao Dai, and Gim~Hee Lee.
\newblock Event-aided dense and continuous point tracking: Everywhere and
  anytime.
\newblock In \emph{Proceedings of the IEEE/CVF International Conference on
  Computer Vision (ICCV)}, 2025.

\bibitem[Wang et~al.(2025)Wang, Zhou, Liu, and Yan]{wang2025diffabflow}
Haonan Wang, Hanyu Zhou, Haoyue Liu, and Luxin Yan.
\newblock Injecting frame-event complementary fusion into diffusion for optical
  flow in challenging scenes.
\newblock In \emph{Advances in Neural Information Processing Systems}, 2025.

\bibitem[Wang et~al.(2024)Wang, Lipson, and Deng]{wang2024searaft}
Yihan Wang, Lahav Lipson, and Jia Deng.
\newblock {SEA-RAFT}: Simple, efficient, accurate {RAFT} for optical flow.
\newblock In \emph{Proceedings of the European Conference on Computer Vision
  (ECCV)}, 2024.

\bibitem[Wu et~al.(2026)Wu, Bai, Yu, Chen, Peng, and Xu]{wu2026ewave}
Jiale Wu, Xiaoyang Bai, Haoming Yu, Yiwei Chen, Yifan Peng, and Weiwei Xu.
\newblock {E-WAVE}: Event-based continuous optical flow via warping-aligned
  visual encoding.
\newblock \emph{arXiv preprint arXiv:2609.34346}, 2026.

\bibitem[Wu et~al.(2024)Wu, Paredes-Vall{\'e}s, and de~Croon]{wu2024idnet}
Yilun Wu, Federico Paredes-Vall{\'e}s, and Guido C. H.~E. de Croon.
\newblock Lightweight event-based optical flow estimation via iterative
  deblurring.
\newblock In \emph{IEEE International Conference on Robotics and Automation
  (ICRA)}, 2024.

\bibitem[Xu et~al.(2026)Xu, Lin, Zhang, Luo, Sun, and Yang]{xu2026bat}
Gangwei Xu, Haotong Lin, Zhaoxing Zhang, Hongcheng Luo, Haiyang Sun, and Xin
  Yang.
\newblock {BAT}: Learning event-based optical flow with bidirectional adaptive
  temporal correlation.
\newblock In \emph{Proceedings of the AAAI Conference on Artificial
  Intelligence}, 2026.

\bibitem[Yang et~al.(2026)Yang, Hong, Son, Lee, Hong, Kim, and
  Kim]{yang2026teto}
Jini Yang, Eunbeen Hong, Soowon Son, Hyunkoo Lee, Sunghwan Hong, Sunok Kim, and
  Seungryong Kim.
\newblock {TETO}: Tracking events with teacher observation for motion
  estimation and frame interpolation.
\newblock \emph{arXiv preprint arXiv:2603.23487}, 2026.

\bibitem[Yang et~al.(2024{\natexlab{a}})Yang, Kang, Huang, Zhao, Xu, Feng, and
  Zhao]{yang2024dav2}
Lihe Yang, Bingyi Kang, Zilong Huang, Zhen Zhao, Xiaogang Xu, Jiashi Feng, and
  Hengshuang Zhao.
\newblock {Depth Anything V2}.
\newblock In \emph{Advances in Neural Information Processing Systems
  (NeurIPS)}, 2024{\natexlab{a}}.

\bibitem[Yang et~al.(2024{\natexlab{b}})Yang, Pan, and Liu]{yang2024ecddp}
Yan Yang, Liyuan Pan, and Liu Liu.
\newblock Event camera data dense pre-training.
\newblock In \emph{Proceedings of the European Conference on Computer Vision
  (ECCV)}, 2024{\natexlab{b}}.

\bibitem[Ye et~al.(2020)Ye, Mitrokhin, Ferm{\"u}ller, Yorke, and
  Aloimonos]{ye2020ecn}
Chengxi Ye, Anton Mitrokhin, Cornelia Ferm{\"u}ller, James~A. Yorke, and
  Yiannis Aloimonos.
\newblock Unsupervised learning of dense optical flow, depth and egomotion with
  event-based sensors.
\newblock In \emph{IEEE/RSJ International Conference on Intelligent Robots and
  Systems (IROS)}, 2020.

\bibitem[Zhang et~al.(2025)Zhang, Zhu, Wang, Wang, and Huang]{zhang2025ematch}
Pengjie Zhang, Lin Zhu, Xiao Wang, Lizhi Wang, and Hua Huang.
\newblock {EMatch}: A unified framework for event-based optical flow and stereo
  matching.
\newblock In \emph{Proceedings of the IEEE/CVF International Conference on
  Computer Vision (ICCV)}, 2025.

\bibitem[Zhou et~al.(2025)Zhou, Wang, Liu, Duan, Chang, and
  Yan]{zhou2025bridge}
Hanyu Zhou, Haonan Wang, Haoyue Liu, Yuxing Duan, Yi Chang, and Luxin Yan.
\newblock Bridge frame and event: Common spatiotemporal fusion for high-dynamic
  scene optical flow.
\newblock In \emph{Proceedings of the IEEE/CVF Conference on Computer Vision
  and Pattern Recognition (CVPR)}, pages 27904--27913, 2025.

\bibitem[Zhou et~al.(2026{\natexlab{a}})Zhou, Hou, Yang, Deng, Li, and
  Xiong]{stflow2025}
Qianang Zhou, Junhui Hou, Meiyi Yang, Yongjian Deng, Youfu Li, and Junlin
  Xiong.
\newblock Spatially-guided temporal aggregation for robust event-{RGB} optical
  flow estimation.
\newblock \emph{IEEE Transactions on Multimedia}, 28:\penalty0 5084--5095,
  2026{\natexlab{a}}.

\bibitem[Zhou et~al.(2026{\natexlab{b}})Zhou, Zhu, Hou, Deng, Li, and
  Xiong]{zhou2026resflow}
Qianang Zhou, Zhiyu Zhu, Junhui Hou, Yongjian Deng, Youfu Li, and Junlin Xiong.
\newblock {ResFlow}: Fine-tuning residual optical flow for event-based high
  temporal resolution motion estimation.
\newblock \emph{IEEE Transactions on Circuits and Systems for Video
  Technology}, 36\penalty0 (3):\penalty0 3527--3538, 2026{\natexlab{b}}.

\bibitem[Zhu et~al.(2018{\natexlab{a}})Zhu, Thakur, {\"O}zaslan, Pfrommer,
  Kumar, and Daniilidis]{zhu2018mvsec}
Alex~Zihao Zhu, Dinesh Thakur, Tolga {\"O}zaslan, Bernd Pfrommer, Vijay Kumar,
  and Kostas Daniilidis.
\newblock The multivehicle stereo event camera dataset: An event camera dataset
  for {3D} perception.
\newblock \emph{IEEE Robotics and Automation Letters}, 2018{\natexlab{a}}.

\bibitem[Zhu et~al.(2018{\natexlab{b}})Zhu, Yuan, Chaney, and
  Daniilidis]{zhu2018evflownet}
Alex~Zihao Zhu, Liangzhe Yuan, Kenneth Chaney, and Kostas Daniilidis.
\newblock {EV-FlowNet}: Self-supervised optical flow estimation for event-based
  cameras.
\newblock In \emph{Robotics: Science and Systems (RSS)}, 2018{\natexlab{b}}.

\bibitem[Zhu et~al.(2019)Zhu, Yuan, Chaney, and
  Daniilidis]{zhu2019unsupervised}
Alex~Zihao Zhu, Liangzhe Yuan, Kenneth Chaney, and Kostas Daniilidis.
\newblock Unsupervised event-based learning of optical flow, depth, and
  egomotion.
\newblock In \emph{Proceedings of the IEEE/CVF Conference on Computer Vision
  and Pattern Recognition (CVPR)}, 2019.

\end{thebibliography}
}

\clearpage
\setcounter{section}{0}
\setcounter{table}{0}
\setcounter{figure}{0}
\setcounter{equation}{0}
\renewcommand{\thesection}{S\arabic{section}}
\renewcommand{\thetable}{S\arabic{table}}
\renewcommand{\thefigure}{S\arabic{figure}}
\renewcommand{\theequation}{S\arabic{equation}}
\renewcommand{\theHsection}{S\arabic{section}}
\renewcommand{\theHtable}{S\arabic{table}}
\renewcommand{\theHfigure}{S\arabic{figure}}
\renewcommand{\theHequation}{S\arabic{equation}}
\renewcommand{\topfraction}{0.9}
\renewcommand{\dbltopfraction}{0.9}
\renewcommand{\bottomfraction}{0.7}
\renewcommand{\textfraction}{0.08}
\renewcommand{\floatpagefraction}{0.8}
\renewcommand{\dblfloatpagefraction}{0.8}
\setcounter{topnumber}{4}
\setcounter{dbltopnumber}{4}
\setcounter{bottomnumber}{2}
\setcounter{totalnumber}{6}
\maketitlesupplementary
In this supplementary material, we provide implementation details, additional comparisons and
additional analyses of \ours{}. In \cref{sec:supp_details}, we report all settings required to
reproduce our results. In \cref{sec:supp_quant}, we compare \ours{} with state-of-the-art methods,
both quantitatively and qualitatively, and show that its improvement stems from dense guidance and
not from the frame camera alone. In \cref{sec:supp_ablation}, we analyze why dense guidance
outperforms fusion at the output and study the robustness of our method to sensor degradation.
Numbers without the prefix S refer to the main paper. We report validation results on the held-out
sequences \texttt{zurich\_city\_05\_a} and \texttt{zurich\_city\_11\_c}, and test results from the
DSEC-Flow benchmark server.

\section{Implementation Details}
\label{sec:supp_details}

\begin{table*}[t]
\centering
\small
\setlength{\tabcolsep}{4pt}
\renewcommand{\arraystretch}{1.06}
\begin{tabular}{@{}p{0.155\linewidth}p{0.815\linewidth}@{}}
\toprule
\textbf{Events} & Two $100$\,ms windows, $B{=}15$ bins each, trilinear splatting, standardized per
grid \\
\textbf{Estimator} & ResNet-18 FPN encoders, hidden dimension $128$, two ConvNeXt update blocks,
$K{=}6$ iterations in training and $4$ at inference, 4-level correlation pyramid with radius $4$,
$18.2$M trainable parameters \\
\textbf{Depth Prior} & Depth AnyEvent ($24.8$M), released DSEC-fine-tuned weights, frozen.
Features projected into the correlation features (Eq.~(2)), eight rigid-motion bases \\
\textbf{Guide} & FlowSeek-T ($42.7$M), released weights, frozen. Both directions on the
$1440{\times}1080$ frame pair, $12$ iterations, mirror-averaged, gate $\tau{=}1.5$\,px (Eq.~(4)),
computed offline \\
\textbf{Guidance Encoder} & Three $3{\times}3$ convolutions at $1/8$ resolution (Eq.~(8)), last
layer zero-initialized, $0.19$M parameters \\
\textbf{Loss} & Laplace-mixture negative log-likelihood ($\gamma{=}0.85$), $\lambda_e{=}1$,
$\lambda_{\text{ss}}{=}5$, $\lambda_{\text{sm}}{=}0.05$, $\lambda_{\text{div}}{=}0.1$ (Eq.~(10)) \\
\textbf{Optimization} & AdamW, weight decay $10^{-4}$, gradient clipping $1.0$, OneCycle schedule
with $5\%$ warm-up, peak learning rate $2{\times}10^{-4}$, batch $6$, bfloat16 \\
\textbf{Schedule} & $60$k steps on MultiFlow (intervals of $20$--$400$\,ms), initialized from the
frame-based FlowSeek-T weights, then $120$k steps on DSEC with guidance ($320{\times}448$ crops that may
overhang the sensor by up to $64$\,px, horizontal flip). Checkpoint with the lowest validation EPE
(step $115$k) \\
\textbf{Inference} & $480{\times}640$, $K{=}4$, four-member symmetric ensemble $\{$native,
$0.75{\times}\}{\times}\{$plain, mirrored$\}$ (Sec.~3.6), epipolar projection (MAGSAC++,
$0.3$\,px threshold, $8{,}000$ points) \\
\textbf{MVSEC} & Fine-tuned for $30$k steps on \texttt{outdoor\_day2} at $dt{\in}\{1,2,3\}$,
$224{\times}288$ crops, batch $8$, peak learning rate $10^{-4}$, horizontal and vertical flips.
Checkpoint chosen on a held-out slice of \texttt{outdoor\_day2} \\
\bottomrule
\end{tabular}
\caption{\textbf{Training and Inference Settings} of the single model of Sec.~3.5.}
\label{tab:details}
\end{table*}

\cref{tab:details} summarizes the training and inference settings of the single model of Sec.~3.5.
Our event-only model, \ours{} (E), serves as the reference throughout this document. It uses the
settings of \cref{tab:details} without the guidance rows, the MultiFlow stage and the inference
row. It has $18.0$M trainable parameters, and we evaluate it in a single pass with $K{=}6$
iterations, without the symmetric ensemble, using its checkpoint at step $75$k.

\paragraph{5-Model Ensemble.} Our final result in Tab.~1 is the average of five guided models with
the architecture of \cref{tab:details}, which we also train on larger displacements. Specifically,
half of their DSEC samples span $200$ or $300$\,ms. We obtain the ground truth of these samples by
chaining consecutive $100$\,ms labels and keep a pixel only if every link is valid. We run the
guide between the two frames at the ends of the span, so that it reads no frame outside the
interval it labels. We train three models independently for $120$k steps from the MultiFlow stage.
We fine-tune the other two from the single model for $30$k steps with a peak learning rate of
$5{\times}10^{-5}$, one of them with the guide inputs (c) below. We compute the guide inputs only
from the frames at the start, middle and end of each $100$\,ms interval, in three variants: (a)
FlowSeek-T on the end frames, (b) a per-pixel choice between (a) and FlowSeek-T chained through
the middle frame, which keeps the hypothesis with the smaller forward--backward residual, and
(c) MEMFOF~\cite{bargatin2025memfof} on the frame triplet. We weight the predictions under (a)
and (b), from four models each, by $3/8$ per set, and those under (c), from all five, by $1/4$.
Each prediction uses the symmetric ensemble and the epipolar projection of \cref{tab:details}, and
the same pipeline and parameters serve every test frame.

\paragraph{Epipolar Projection.} The ground truth of DSEC is the flow of the static scene under
ego-motion. Indeed, an essential matrix fitted to the validation ground truth leaves a mean
epipolar residual of only $0.002$\,px. This motivates the per-frame projection of Sec.~3.6, for
which we do not tune any parameter on the test set.

\paragraph{Depth-Prior Ablation.} The ablation of Sec.~4.4 retrains an earlier event-only
recipe ($30$k steps, $288{\times}384$ crops, $K{=}4$) with the inputs of the prior set to zero.

\section{Comparison with the State of the Art}
\label{sec:supp_quant}

\begin{table*}[t]
\centering
\footnotesize
\setlength{\tabcolsep}{2.4pt}
\resizebox{\ifdim\width>\textwidth\textwidth\else\width\fi}{!}{%
\begin{tabular}{ll ccc ccc ccc ccc ccc ccc ccc ccc}
\toprule
& & \multicolumn{3}{c}{Overall} & \multicolumn{3}{c}{\texttt{int\_00\_b}} & \multicolumn{3}{c}{\texttt{int\_01\_a}} & \multicolumn{3}{c}{\texttt{thun\_01\_a}} & \multicolumn{3}{c}{\texttt{thun\_01\_b}} & \multicolumn{3}{c}{\texttt{zur\_12\_a}} & \multicolumn{3}{c}{\texttt{zur\_14\_c}} & \multicolumn{3}{c}{\texttt{zur\_15\_a}} \\
\cmidrule(lr){3-5}\cmidrule(lr){6-8}\cmidrule(lr){9-11}\cmidrule(lr){12-14}\cmidrule(lr){15-17}\cmidrule(lr){18-20}\cmidrule(lr){21-23}\cmidrule(lr){24-26}
Method & In & EPE & 3PE & AE & EPE & 3PE & AE & EPE & 3PE & AE & EPE & 3PE & AE & EPE & 3PE & AE & EPE & 3PE & AE & EPE & 3PE & AE & EPE & 3PE & AE \\
\midrule
E-RAFT~\cite{gehrig2021eraft} & E & 0.79 & 2.68 & 2.85 & 1.39 & 6.19 & 2.36 & 0.90 & 3.91 & 2.54 & 0.65 & 1.87 & 2.94 & 0.58 & 1.52 & 2.20 & 0.61 & 1.06 & 4.50 & 0.71 & 1.91 & 3.43 & 0.59 & 1.30 & 2.55 \\
TMA~\cite{liu2023tma} & E & 0.74 & 2.30 & 2.68 & 1.39 & 5.79 & 2.16 & 0.81 & 3.11 & 2.23 & 0.62 & 1.61 & 2.88 & 0.55 & 1.31 & 2.10 & 0.57 & 0.87 & 4.38 & 0.66 & 1.99 & 3.09 & 0.55 & 1.08 & 2.51 \\
IDNet~\cite{wu2024idnet} & E & 0.72 & 2.04 & 2.72 & 1.25 & 4.35 & 2.11 & 0.77 & 2.60 & 2.25 & 0.57 & 1.47 & 2.66 & 0.55 & 1.35 & 2.07 & 0.60 & 1.16 & 4.56 & 0.76 & 2.74 & 3.74 & 0.55 & 1.02 & 2.55 \\
ECDDP$^{\dagger}$~\cite{yang2024ecddp} & E & 0.70 & 1.96 & 2.58 & 1.31 & 5.03 & 2.00 & 0.76 & 2.47 & 2.18 & 0.52 & 1.36 & 2.34 & 0.51 & 1.21 & 1.93 & 0.55 & 0.76 & 4.35 & 0.69 & 2.39 & 3.22 & 0.52 & 0.89 & 2.41 \\
BFlow~\cite{gehrig2024dense} & E+I & 0.69 & 2.02 & 2.42 & 1.11 & 4.58 & 1.93 & 0.77 & 2.63 & 2.07 & 0.59 & 1.59 & 2.71 & 0.55 & 1.42 & 1.84 & 0.54 & 0.72 & 3.86 & 0.59 & 1.37 & 2.44 & 0.57 & 1.24 & 2.38 \\
STFlow~\cite{stflow2025} & E+I & 0.66 & 1.68 & 2.37 & 1.22 & 4.32 & 1.96 & 0.72 & 2.10 & 2.06 & 0.55 & 1.31 & 2.43 & 0.50 & 1.12 & 1.76 & 0.50 & 0.54 & 3.80 & 0.55 & \textbf{0.94} & 2.47 & 0.53 & 0.92 & 2.27 \\
STFlow$^{\dagger}$~\cite{stflow2025} & E+I & 0.63 & 1.45 & 2.29 & 1.11 & 3.74 & 1.81 & 0.68 & 1.70 & 1.90 & 0.54 & 1.23 & 2.51 & 0.49 & 1.04 & 1.76 & 0.49 & 0.53 & 3.61 & 0.52 & 1.04 & 2.57 & 0.51 & 0.77 & 2.31 \\
E-WAVE$^{\dagger}$~\cite{wu2026ewave} & E+I & 0.62 & 1.56 & 2.24 & 1.09 & 4.22 & 1.67 & 0.66 & 1.79 & 1.74 & 0.57 & 1.21 & 2.83 & 0.50 & 1.10 & 1.90 & \textbf{0.48} & \textbf{0.50} & 3.61 & 0.53 & 0.96 & 2.25 & 0.50 & 0.87 & 2.24 \\
EDCPT$^{\dagger}$~\cite{wan2025edcpt} & E+I & 0.63 & 1.52 & 2.17 & 1.20 & 4.18 & 1.61 & 0.63 & 1.76 & 1.67 & \textbf{0.49} & \textbf{1.06} & 2.40 & \textbf{0.46} & \textbf{1.03} & 1.71 & 0.51 & 0.51 & 3.58 & 0.61 & 0.99 & 2.54 & 0.50 & 0.84 & 2.24 \\
\midrule
\ours{} (E) & E & 0.76 & 2.44 & 2.60 & 1.36 & 5.50 & 2.08 & 0.85 & 3.38 & 2.27 & 0.61 & 1.86 & 2.73 & 0.58 & 1.65 & 2.03 & 0.57 & 0.87 & 4.10 & 0.67 & 1.66 & 2.93 & 0.58 & 1.32 & 2.49 \\
\textbf{\ours}$^{\dagger}$ (E+I) & E+I & \textbf{0.55} & \textbf{1.11} & \textbf{2.08} & \textbf{0.91} & \textbf{2.87} & \textbf{1.53} & \textbf{0.55} & \textbf{0.88} & \textbf{1.60} & \textbf{0.49} & 1.24 & \textbf{2.27} & 0.48 & 1.04 & \textbf{1.65} & \textbf{0.48} & 0.53 & \textbf{3.54} & \textbf{0.48} & 1.08 & \textbf{2.14} & \textbf{0.47} & \textbf{0.68} & \textbf{2.11} \\
\bottomrule
\end{tabular}}
\caption{\textbf{Per-Sequence DSEC-Flow Test Results} (EPE, 3PE and AE from the benchmark server).
BFlow and STFlow rows as tabulated by STFlow~\cite{stflow2025}. $^{\dagger}$Pretrained on synthetic
event data. Our E+I row is the 5-model ensemble of Tab.~1. Best in bold.}
\label{tab:perseqfull}
\end{table*}

\paragraph{Per-Sequence Results.} \cref{tab:perseqfull} reports the result of Tab.~1 for each test
sequence, together with the angular error. \ours{} achieves the lowest or tied-lowest EPE on six of
the seven sequences, as stated in Sec.~4.1, and the lowest AE on all seven. The improvement is
largest on the sequences with the fastest motion. On \texttt{interlaken\_00\_b}, \ours{} reaches
$0.91$\,px, whereas the best published method reaches $1.09$\,px. On \texttt{interlaken\_01\_a},
it reaches $0.55$\,px against $0.63$\,px. This is the regime predicted by the refinement basin of
Sec.~4.4, in which an event-only start is far from the solution. On the night sequence
\texttt{zurich\_city\_12\_a}, where the gate rejects much of the guide, \ours{} still matches the
best published EPE ($0.48$).

\paragraph{Frame-Only Methods.} We now ask whether the improvement of \ours{} stems from guidance or
simply from the frame camera. To this end, we evaluate three frame-only methods,
RAFT~\cite{teed2020raft}, FlowSeek-T~\cite{poggi2025flowseek} and MEMFOF~\cite{bargatin2025memfof},
with the same registration as our guide. None of them was trained on DSEC. On validation, the three methods lie within
$0.011$\,px of each other ($0.527$--$0.538$). They are barely better than our event-only model
($0.546$) and far from the guided model ($0.435$). On the test server, where we submitted them as
declared ablations, they score between $0.692$ and $1.032$. FlowSeek-T, our guide, is the weakest
of the three and reaches $2.26$\,px on the night sequence. Nevertheless, the model it guides
reaches $0.636$ in a single pass at step $35$k and $0.583$ as the single model with the test-time
steps of Sec.~3.6 (Tab.~2). We conclude that
the validation margin of $17$--$19\%$ over the frame methods is the contribution of guidance and
not of the frame camera. A stronger frame network would not close this gap either. Indeed, MEMFOF
improves on FlowSeek-T by only $0.004$\,px over all pixels and not at all on the pixels that the
forward--backward gate of each method accepts ($0.494$ against $0.484$).

\begin{table}[b]
\centering
\footnotesize
\setlength{\tabcolsep}{2.4pt}
\resizebox{\linewidth}{!}{%
\begin{tabular}{l c c ccc cc}
\toprule
 & & Test & \multicolumn{3}{c}{Boundary F1$\uparrow$} & \multicolumn{2}{c}{$d{=}2$} \\
\cmidrule(lr){4-6}\cmidrule(lr){7-8}
Method & In & EPE$\downarrow$ & $d{=}2$ & $d{=}4$ & $d{=}8$ & B-EPE$\downarrow$ & Contr.$\uparrow$ \\
\midrule
E-RAFT~\cite{gehrig2021eraft} & E & 0.788 & 0.22 & 0.41 & 0.66 & 2.12 & 0.26 \\
IDNet~\cite{wu2024idnet} & E & 0.719 & 0.24 & 0.44 & 0.69 & 2.07 & 0.29 \\
BFlow~\cite{gehrig2024dense} & E & 0.75 & 0.19 & 0.38 & 0.63 & 2.32 & 0.23 \\
BFlow~\cite{gehrig2024dense} & E+I & 0.69 & 0.21 & 0.41 & 0.66 & 2.21 & 0.26 \\
\midrule
FlowSeek-T, warped (guide) & I & 1.032 & 0.14 & 0.35 & 0.65 & 2.35 & 0.23 \\
\ours{} (E) & E & 0.759 & 0.18 & 0.36 & 0.63 & 2.28 & 0.23 \\
\textbf{\ours} (E+I) & E+I & \textbf{0.583} & \textbf{0.27} & \textbf{0.49} & \textbf{0.72} & \textbf{2.01} & \textbf{0.33} \\
\bottomrule
\end{tabular}}
\caption{\textbf{Motion Boundaries} on held-out validation ($724$ frames, single pass). F1 at
pixel-pair distance $d$, averaged over flow-difference thresholds of $1$ to $4$\,px. B-EPE: EPE at
the ground-truth boundaries. Contr.: predicted over true flow jump. Test EPE of \ours{} (E+I): single
model. The public checkpoints were trained on these sequences.}
\label{tab:boundary}
\end{table}

\begin{figure*}[t]
\centering
\includegraphics[width=0.9\textwidth]{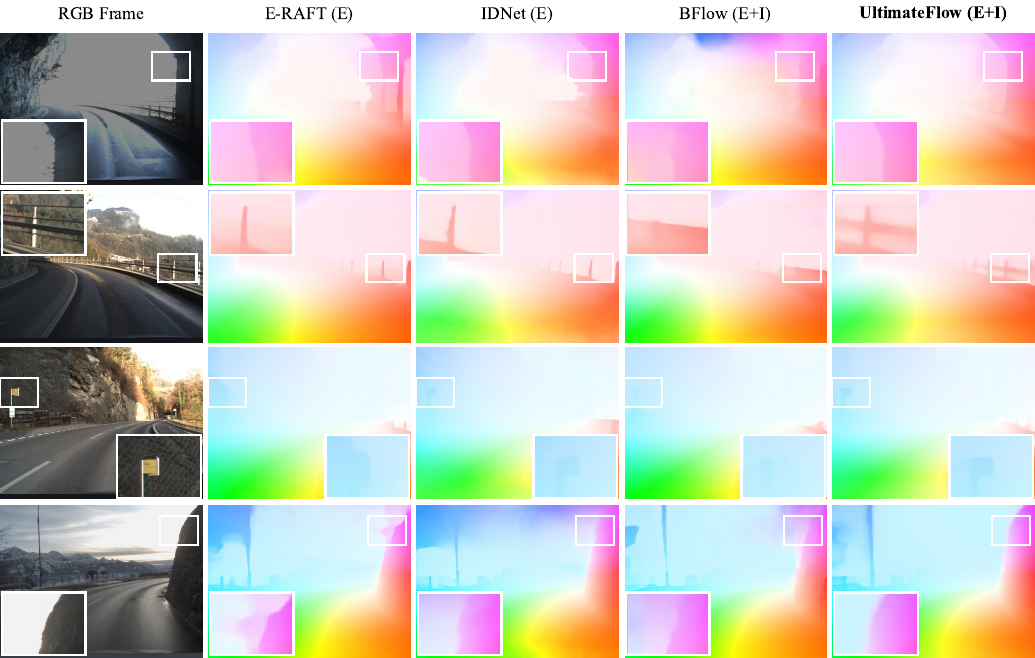}
\caption{\textbf{Further Comparison on the DSEC Test Sequences}. Each box is enlarged in the
inset of the same tile. The tunnel portal, the railing, the bus-stop sign and the rock wall are
sharpest in \ours{} (E+I), the 5-model ensemble. Baselines use their public checkpoints. Dark frames are brightened for
display.}
\label{fig:qual_test}
\end{figure*}

\paragraph{Motion Boundaries.} Frames resolve object outlines that events tend to blur. Therefore, we
expect guidance to help most at motion boundaries. We measure boundary quality with the
occluding-contour F1 of Depth Pro~\cite{bochkovskii2025depthpro} and run every method on identical
inputs (\cref{tab:boundary}). \ours{} achieves the highest F1 at every pair distance, the lowest
boundary EPE and the largest contrast. Note that this split favors the published checkpoints, which
were trained on it. Moreover, the guided estimate is sharper than both of its sources. The warped
guide has the weakest boundaries at $d{=}2$ ($0.14$) because the registration parallax of Prop.~1
shifts them, and the event-only model follows ($0.18$). This shows that guidance creates the
sharpness and does not copy it from the frames, i.e., the guide supplies the outline and the
events place it.

\begin{figure*}[t]
\centering
\includegraphics[width=0.9\textwidth]{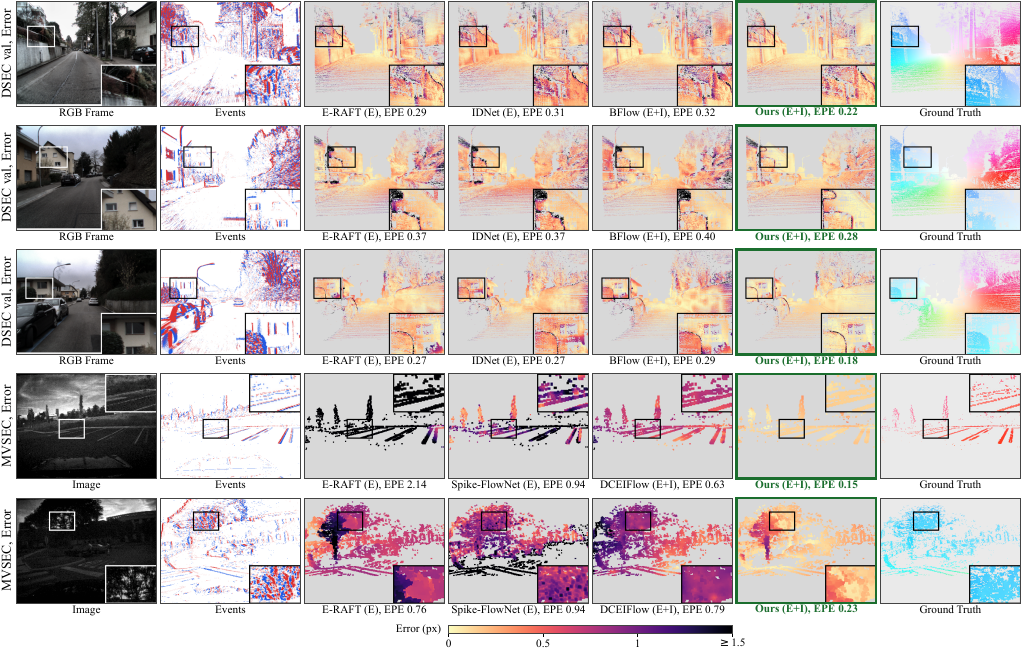}
\caption{\textbf{Error Against Ground Truth} on DSEC validation and MVSEC
\texttt{outdoor\_day1}. Tiles show the per-pixel error on labeled pixels with the EPE of the frame
below, and the last column shows the ground-truth flow. The box marks the region where our error
improves most. The published DSEC checkpoints were trained on these validation sequences. MVSEC
labels are dilated by one pixel for display.}
\label{fig:qual_gt}
\end{figure*}

\paragraph{Qualitative Results.} The main paper compares the methods on the day and night test
sequences (Fig.~4). In \cref{fig:qual_test}, we show four additional test frames from two
sequences. The first frame shows the exit of a tunnel on \texttt{interlaken\_00\_b}. Here, the
frame is overexposed, which is the situation in which a frame-based guide should fail and events
should not. The gate detects this. Its coverage drops from $77\%$ inside the tunnel to $25\%$
under the glare and is $45\%$ on the frame shown. Despite this, \ours{} draws the portal edge more
sharply than every baseline. Once the frame recovers, \ours{} resolves both the post and the rails
of the railing. The remaining two rows show a small and a large object. \ours{} outlines the
bus-stop sign and its pole most completely, and it is the only method with a sharp edge between
the rock wall and the sky at dusk. In \cref{fig:qual_gt}, we show the per-pixel error of each
method on frames with ground truth from DSEC and MVSEC. On DSEC, our error decreases most on near
structures, such as building facades and parked cars. On MVSEC, it decreases across the whole
frame.

\section{Analysis of Dense Guidance}
\label{sec:supp_ablation}
In this section, we analyze the guided estimator. We compare it with output-level fusion and
evaluate it under sensor degradation.

\begin{figure}[t]
\centering
\includegraphics[width=\linewidth]{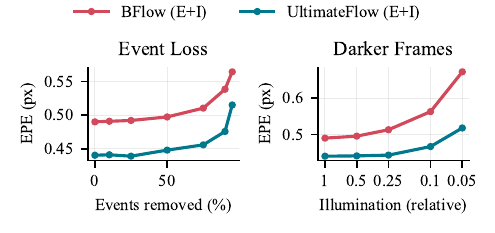}
\caption{\textbf{EPE Under Increasing Degradation} of the events (left) and of the frame
illumination (right).}
\label{fig:robust_curves}
\end{figure}

\begin{table}[t]
\centering
\small
\setlength{\tabcolsep}{4pt}
\resizebox{\linewidth}{!}{%
\begin{tabular}{lcc}
\toprule
Rule (Validation, $111.7$M Labeled px) & EPE$\downarrow$ & vs.\ \ours{} (E) \\
\midrule
\ours{} (E) & $0.546$ & -- \\
FlowSeek-T guide, gate-accepted pixels ($97\%$ of labeled) & $0.484$ & -- \\
\midrule
Output blend, averaging ($w{=}0.5$) & $0.487$ & $-10.7\%$ \\
Output blend, server-selected $w{=}0.70$ & $0.488$ & $-10.5\%$ \\
Output blend, initial $w{=}0.85$ (Eq.~(7)) & $0.499$ & $-8.6\%$ \\
Output blend, substitution ($w{=}1$) & $0.520$ & $-4.7\%$ \\
\midrule
Dense guidance (Sec.~3.4) & $\mathbf{0.435}$ & $\mathbf{-20.3\%}$ \\
Dense guidance $+$ output blend ($w{=}0.70$) & $0.470$ & $-13.8\%$ \\
\midrule
Per-pixel oracle (uses ground truth) & $0.388$ & $-28.8\%$ \\
\bottomrule
\end{tabular}}
\caption{\textbf{Output-Level Fusion Rules Versus Dense Guidance} on held-out validation,
single pass, $K{=}6$. Output rules keep the event-only estimate on gate-rejected pixels ($3\%$ of
labeled pixels). Best result without ground truth in bold.}
\label{tab:fusionrules}
\end{table}

\paragraph{Dense Guidance Versus Output Fusion.} A common way to combine two flow estimates is to
blend them at the output (Eq.~(7)). In \cref{tab:fusionrules}, we evaluate all such rules under the
most favorable conditions, including the weight selected on the test server. None of them comes
close to dense guidance. Specifically, the best blend improves on the event-only model by
$10.7\%$, whereas dense guidance improves on it by $20.3\%$. Dense guidance thereby closes $53\%$
of the gap between the best blend and a per-pixel oracle that knows the ground truth. Blending the
guide once more into the guided estimate even degrades it ($0.470$), which Tab.~2 confirms on the
test server. We conclude that the guide is more useful as evidence inside the estimator than as a
second estimate to average with.

\begin{table}[t]
\centering
\small
\setlength{\tabcolsep}{3pt}
\resizebox{\linewidth}{!}{%
\begin{tabular}{llll}
\toprule
Condition (363 Frames) & \ours{} (E) & BFlow (E+I) & \ours{} (E+I) \\
\midrule
Reference & $0.547$ & $0.490$ & $\textbf{0.441}$ \\
Events removed $50\%$ & $0.591$ {\scriptsize$(+8\%)$} & $0.497$ {\scriptsize$(+1\%)$} & $\textbf{0.448}$ {\scriptsize$(+2\%)$} \\
Events removed $90\%$ & $0.907$ {\scriptsize$(+66\%)$} & $0.538$ {\scriptsize$(+10\%)$} & $\textbf{0.476}$ {\scriptsize$(+8\%)$} \\
Events removed $95\%$ & $1.227$ {\scriptsize$(+124\%)$} & $0.564$ {\scriptsize$(+15\%)$} & $\textbf{0.515}$ {\scriptsize$(+17\%)$} \\
Spurious events $+100\%$ & $0.770$ {\scriptsize$(+41\%)$} & $0.512$ {\scriptsize$(+5\%)$} & $\textbf{0.487}$ {\scriptsize$(+10\%)$} \\
\midrule
Illumination $\times0.1$ & -- & $0.563$ {\scriptsize$(+15\%)$} & $\textbf{0.467}$ {\scriptsize$(+6\%)$} \\
Illumination $\times0.05$ & -- & $0.672$ {\scriptsize$(+37\%)$} & $\textbf{0.518}$ {\scriptsize$(+18\%)$} \\
Gamma $3$ & -- & $0.614$ {\scriptsize$(+25\%)$} & $\textbf{0.534}$ {\scriptsize$(+21\%)$} \\
Sensor noise $\sigma{=}8$ & -- & $0.533$ {\scriptsize$(+9\%)$} & $\textbf{0.481}$ {\scriptsize$(+9\%)$} \\
\midrule
Illum.\ $\times0.05$, events $-90\%$ & -- & $0.803$ {\scriptsize$(+64\%)$} & $\textbf{0.603}$ {\scriptsize$(+37\%)$} \\
\bottomrule
\end{tabular}}
\caption{\textbf{Robustness to Sensor Degradation} against BFlow~\cite{gehrig2024dense}
(E+I). Validation EPE on every fourth frame, single pass, with the change from each method's own
reference in parentheses. Events are removed or added uniformly at random. Frame conditions degrade
the RGB frames read by BFlow and by our guide, with the gate recomputed. BFlow was trained on these
sequences. Best E+I method in bold.}
\label{tab:robust}
\end{table}

\begin{figure*}[t]
\centering
\includegraphics[width=0.8\textwidth]{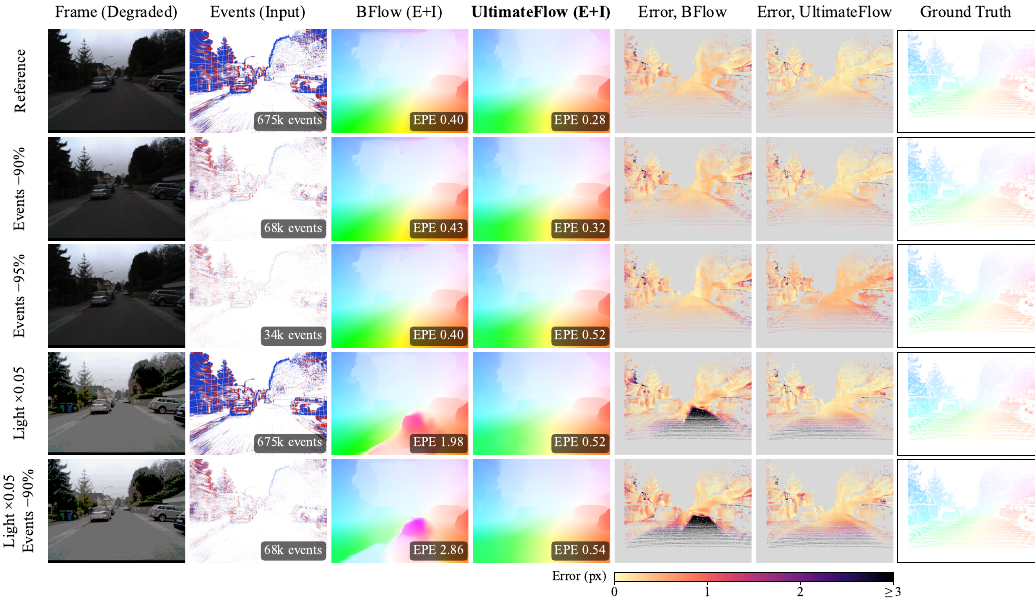}
\caption{\textbf{One Validation Frame Under Degradation} (\texttt{zurich\_city\_11\_c}),
single pass. The second column shows the events that both methods read. Darkened frames are
rescaled for display. Error maps use the color scale at the bottom, and gray marks pixels without ground
truth. Badges give the number of events or the EPE. With $95\%$ of the events removed, BFlow is
ahead on this frame, whereas \ours{} is ahead on average (\cref{tab:robust}).}
\label{fig:robust_qual}
\end{figure*}

\paragraph{Robustness to Sensor Degradation.} Sec.~4.6 summarizes how the guided model behaves when a
sensor degrades. Here, we compare it with BFlow (E+I), which reads the same events and frames, at
the same degradation levels (\cref{tab:robust} and \cref{fig:robust_curves}). \ours{} (E+I)
achieves the lower EPE under every condition of \cref{tab:robust}. Note that BFlow was trained on
these sequences. The difference is largest
when the frames darken. With $5\%$ of the light, the EPE of BFlow increases by $37\%$ and that of
\ours{} by $18\%$. When we additionally remove $90\%$ of the events, the increases are $64\%$ and
$37\%$. The reason is that a dark but noise-free frame still yields a self-consistent guide, which
the gate keeps ($75\%$ coverage at ${\times}0.05$), whereas BFlow reads the dark frame directly.
Event loss affects both methods mildly, by $15$ to $17\%$ at $95\%$ removal. Even then, \ours{}
(E+I) remains more accurate than our event-only model on the intact stream ($0.515$ against
$0.547$\,px), because the flow state initialized from the guide carries the estimate where the
correlation volume has lost its evidence. Under mild sensor noise ($\sigma{=}8$), the gate still
accepts $74\%$ of the pixels, and both methods degrade by $9\%$. Stronger frame noise collapses the
gate, which is the limitation stated in Sec.~5. \cref{fig:robust_qual}
illustrates these findings on one frame. With $90\%$ of the events removed, the input retains only
sparse edges, and both flows stay close to the reference. With $95\%$ removed, \ours{} degrades
more than BFlow on this frame, although its EPE is lower on average. Under darkening, BFlow loses
the flow of the road ahead, whereas \ours{} preserves it.

\end{document}